\documentclass[twoside,11pt]{article}

\usepackage{bbm}
\usepackage{algorithmicx}
\usepackage{jmlr2e}
\usepackage{amsmath}
\usepackage{xcolor}
\usepackage{bm}
\usepackage{tikz}
\usetikzlibrary{bayesnet}
\usepackage{subcaption}
\usepackage{float}
\usepackage{stackengine}
\usepackage{array}
\usepackage{multirow}
\usepackage{stmaryrd}
\usepackage{pgfplots}
\usepgfplotslibrary{fillbetween}
\usetikzlibrary{patterns}
\usetikzlibrary{shapes.geometric}
\usepackage[ruled,vlined]{algorithm2e}
\usepackage{xstring}
\usepackage[margin=1cm,bottom=1in]{geometry}
\usepackage{color,fancyvrb}

\let\oldmathcal\mathcal
\DeclareMathAlphabet{\mathdutchcal}{U}{dutchcal}{m}{n}

\renewcommand{\mathcal}[1]{%
  \IfSubStr{ABCDEFGHIJKLMNOPQRSTUVWXYZ}{#1}{\oldmathcal{#1}}{\mathdutchcal{#1}}
}
\makeatother

\pgfmathdeclarefunction{gauss}{3}{%
  \pgfmathparse{1/(#3*sqrt(2*pi))*exp(-((#1-#2)^2)/(2*#3^2))}%
}
\pgfmathdeclarefunction{sum_gauss}{5}{%
  \pgfmathparse{1/(3*#3*sqrt(2*pi))*exp(-((#1-#2)^2)/(2*#3^2)) + 2/(3*#5*sqrt(2*pi))*exp(-((#1-#4)^2)/(2*#5^2))}%
}

\DeclareMathOperator*{\argmax}{arg\,max}

\definecolor{blue-violet}{rgb}{0.54, 0.17, 0.89}
\definecolor{bluepigment}{rgb}{0.2, 0.2, 0.6}
\makeatletter
\newcommand*\bigcdot{\mathpalette\bigcdot@{.5}}
\newcommand*\bigcdot@[2]{\mathbin{\vcenter{\hbox{\scalebox{#2}{$\m@th#1\bullet$}}}}}
\makeatother
\def\delequal{\mathrel{\ensurestackMath{\stackon[1pt]{=}{\scriptstyle\Delta}}}}
\definecolor{Yellow}{RGB}{211, 176, 15}
\definecolor{Green}{rgb}{0.01, 0.75, 0.24}
\colorlet{Red}{red!50!black}
\colorlet{Blue}{blue!50!black}
\definecolor{Violet}{rgb}{0.56,0.14,0.56}
\newcommand{\indep}{\perp \! \! \! \perp}

\newcommand{\heart}{\ensuremath\heartsuit}

\newcommand{\nb}[1]{\# #1}

\makeatletter
\def\PYGdefault@reset{\let\PYGdefault@it=\relax \let\PYGdefault@bf=\relax%
    \let\PYGdefault@ul=\relax \let\PYGdefault@tc=\relax%
    \let\PYGdefault@bc=\relax \let\PYGdefault@ff=\relax}
\def\PYGdefault@tok#1{\csname PYGdefault@tok@#1\endcsname}
\def\PYGdefault@toks#1+{\ifx\relax#1\empty\else%
    \PYGdefault@tok{#1}\expandafter\PYGdefault@toks\fi}
\def\PYGdefault@do#1{\PYGdefault@bc{\PYGdefault@tc{\PYGdefault@ul{%
    \PYGdefault@it{\PYGdefault@bf{\PYGdefault@ff{#1}}}}}}}
\def\PYGdefault#1#2{\PYGdefault@reset\PYGdefault@toks#1+\relax+\PYGdefault@do{#2}}

\expandafter\def\csname PYGdefault@tok@w\endcsname{\def\PYGdefault@tc##1{\textcolor[rgb]{0.73,0.73,0.73}{##1}}}
\expandafter\def\csname PYGdefault@tok@c\endcsname{\let\PYGdefault@it=\textit\def\PYGdefault@tc##1{\textcolor[rgb]{0.25,0.50,0.50}{##1}}}
\expandafter\def\csname PYGdefault@tok@cp\endcsname{\def\PYGdefault@tc##1{\textcolor[rgb]{0.74,0.48,0.00}{##1}}}
\expandafter\def\csname PYGdefault@tok@k\endcsname{\let\PYGdefault@bf=\textbf\def\PYGdefault@tc##1{\textcolor[rgb]{0.00,0.50,0.00}{##1}}}
\expandafter\def\csname PYGdefault@tok@kp\endcsname{\def\PYGdefault@tc##1{\textcolor[rgb]{0.00,0.50,0.00}{##1}}}
\expandafter\def\csname PYGdefault@tok@kt\endcsname{\def\PYGdefault@tc##1{\textcolor[rgb]{0.69,0.00,0.25}{##1}}}
\expandafter\def\csname PYGdefault@tok@o\endcsname{\def\PYGdefault@tc##1{\textcolor[rgb]{0.40,0.40,0.40}{##1}}}
\expandafter\def\csname PYGdefault@tok@ow\endcsname{\let\PYGdefault@bf=\textbf\def\PYGdefault@tc##1{\textcolor[rgb]{0.67,0.13,1.00}{##1}}}
\expandafter\def\csname PYGdefault@tok@nb\endcsname{\def\PYGdefault@tc##1{\textcolor[rgb]{0.00,0.50,0.00}{##1}}}
\expandafter\def\csname PYGdefault@tok@nf\endcsname{\def\PYGdefault@tc##1{\textcolor[rgb]{0.00,0.00,1.00}{##1}}}
\expandafter\def\csname PYGdefault@tok@nc\endcsname{\let\PYGdefault@bf=\textbf\def\PYGdefault@tc##1{\textcolor[rgb]{0.00,0.00,1.00}{##1}}}
\expandafter\def\csname PYGdefault@tok@nn\endcsname{\let\PYGdefault@bf=\textbf\def\PYGdefault@tc##1{\textcolor[rgb]{0.00,0.00,1.00}{##1}}}
\expandafter\def\csname PYGdefault@tok@ne\endcsname{\let\PYGdefault@bf=\textbf\def\PYGdefault@tc##1{\textcolor[rgb]{0.82,0.25,0.23}{##1}}}
\expandafter\def\csname PYGdefault@tok@nv\endcsname{\def\PYGdefault@tc##1{\textcolor[rgb]{0.10,0.09,0.49}{##1}}}
\expandafter\def\csname PYGdefault@tok@no\endcsname{\def\PYGdefault@tc##1{\textcolor[rgb]{0.53,0.00,0.00}{##1}}}
\expandafter\def\csname PYGdefault@tok@nl\endcsname{\def\PYGdefault@tc##1{\textcolor[rgb]{0.63,0.63,0.00}{##1}}}
\expandafter\def\csname PYGdefault@tok@ni\endcsname{\let\PYGdefault@bf=\textbf\def\PYGdefault@tc##1{\textcolor[rgb]{0.60,0.60,0.60}{##1}}}
\expandafter\def\csname PYGdefault@tok@na\endcsname{\def\PYGdefault@tc##1{\textcolor[rgb]{0.49,0.56,0.16}{##1}}}
\expandafter\def\csname PYGdefault@tok@nt\endcsname{\let\PYGdefault@bf=\textbf\def\PYGdefault@tc##1{\textcolor[rgb]{0.00,0.50,0.00}{##1}}}
\expandafter\def\csname PYGdefault@tok@nd\endcsname{\def\PYGdefault@tc##1{\textcolor[rgb]{0.67,0.13,1.00}{##1}}}
\expandafter\def\csname PYGdefault@tok@s\endcsname{\def\PYGdefault@tc##1{\textcolor[rgb]{0.73,0.13,0.13}{##1}}}
\expandafter\def\csname PYGdefault@tok@sd\endcsname{\let\PYGdefault@it=\textit\def\PYGdefault@tc##1{\textcolor[rgb]{0.73,0.13,0.13}{##1}}}
\expandafter\def\csname PYGdefault@tok@si\endcsname{\let\PYGdefault@bf=\textbf\def\PYGdefault@tc##1{\textcolor[rgb]{0.73,0.40,0.53}{##1}}}
\expandafter\def\csname PYGdefault@tok@se\endcsname{\let\PYGdefault@bf=\textbf\def\PYGdefault@tc##1{\textcolor[rgb]{0.73,0.40,0.13}{##1}}}
\expandafter\def\csname PYGdefault@tok@sr\endcsname{\def\PYGdefault@tc##1{\textcolor[rgb]{0.73,0.40,0.53}{##1}}}
\expandafter\def\csname PYGdefault@tok@ss\endcsname{\def\PYGdefault@tc##1{\textcolor[rgb]{0.10,0.09,0.49}{##1}}}
\expandafter\def\csname PYGdefault@tok@sx\endcsname{\def\PYGdefault@tc##1{\textcolor[rgb]{0.00,0.50,0.00}{##1}}}
\expandafter\def\csname PYGdefault@tok@m\endcsname{\def\PYGdefault@tc##1{\textcolor[rgb]{0.40,0.40,0.40}{##1}}}
\expandafter\def\csname PYGdefault@tok@gh\endcsname{\let\PYGdefault@bf=\textbf\def\PYGdefault@tc##1{\textcolor[rgb]{0.00,0.00,0.50}{##1}}}
\expandafter\def\csname PYGdefault@tok@gu\endcsname{\let\PYGdefault@bf=\textbf\def\PYGdefault@tc##1{\textcolor[rgb]{0.50,0.00,0.50}{##1}}}
\expandafter\def\csname PYGdefault@tok@gd\endcsname{\def\PYGdefault@tc##1{\textcolor[rgb]{0.63,0.00,0.00}{##1}}}
\expandafter\def\csname PYGdefault@tok@gi\endcsname{\def\PYGdefault@tc##1{\textcolor[rgb]{0.00,0.63,0.00}{##1}}}
\expandafter\def\csname PYGdefault@tok@gr\endcsname{\def\PYGdefault@tc##1{\textcolor[rgb]{1.00,0.00,0.00}{##1}}}
\expandafter\def\csname PYGdefault@tok@ge\endcsname{\let\PYGdefault@it=\textit}
\expandafter\def\csname PYGdefault@tok@gs\endcsname{\let\PYGdefault@bf=\textbf}
\expandafter\def\csname PYGdefault@tok@gp\endcsname{\let\PYGdefault@bf=\textbf\def\PYGdefault@tc##1{\textcolor[rgb]{0.00,0.00,0.50}{##1}}}
\expandafter\def\csname PYGdefault@tok@go\endcsname{\def\PYGdefault@tc##1{\textcolor[rgb]{0.53,0.53,0.53}{##1}}}
\expandafter\def\csname PYGdefault@tok@gt\endcsname{\def\PYGdefault@tc##1{\textcolor[rgb]{0.00,0.27,0.87}{##1}}}
\expandafter\def\csname PYGdefault@tok@err\endcsname{\def\PYGdefault@bc##1{\setlength{\fboxsep}{0pt}\fcolorbox[rgb]{1.00,0.00,0.00}{1,1,1}{\strut ##1}}}
\expandafter\def\csname PYGdefault@tok@kc\endcsname{\let\PYGdefault@bf=\textbf\def\PYGdefault@tc##1{\textcolor[rgb]{0.00,0.50,0.00}{##1}}}
\expandafter\def\csname PYGdefault@tok@kd\endcsname{\let\PYGdefault@bf=\textbf\def\PYGdefault@tc##1{\textcolor[rgb]{0.00,0.50,0.00}{##1}}}
\expandafter\def\csname PYGdefault@tok@kn\endcsname{\let\PYGdefault@bf=\textbf\def\PYGdefault@tc##1{\textcolor[rgb]{0.00,0.50,0.00}{##1}}}
\expandafter\def\csname PYGdefault@tok@kr\endcsname{\let\PYGdefault@bf=\textbf\def\PYGdefault@tc##1{\textcolor[rgb]{0.00,0.50,0.00}{##1}}}
\expandafter\def\csname PYGdefault@tok@bp\endcsname{\def\PYGdefault@tc##1{\textcolor[rgb]{0.00,0.50,0.00}{##1}}}
\expandafter\def\csname PYGdefault@tok@fm\endcsname{\def\PYGdefault@tc##1{\textcolor[rgb]{0.00,0.00,1.00}{##1}}}
\expandafter\def\csname PYGdefault@tok@vc\endcsname{\def\PYGdefault@tc##1{\textcolor[rgb]{0.10,0.09,0.49}{##1}}}
\expandafter\def\csname PYGdefault@tok@vg\endcsname{\def\PYGdefault@tc##1{\textcolor[rgb]{0.10,0.09,0.49}{##1}}}
\expandafter\def\csname PYGdefault@tok@vi\endcsname{\def\PYGdefault@tc##1{\textcolor[rgb]{0.10,0.09,0.49}{##1}}}
\expandafter\def\csname PYGdefault@tok@vm\endcsname{\def\PYGdefault@tc##1{\textcolor[rgb]{0.10,0.09,0.49}{##1}}}
\expandafter\def\csname PYGdefault@tok@sa\endcsname{\def\PYGdefault@tc##1{\textcolor[rgb]{0.73,0.13,0.13}{##1}}}
\expandafter\def\csname PYGdefault@tok@sb\endcsname{\def\PYGdefault@tc##1{\textcolor[rgb]{0.73,0.13,0.13}{##1}}}
\expandafter\def\csname PYGdefault@tok@sc\endcsname{\def\PYGdefault@tc##1{\textcolor[rgb]{0.73,0.13,0.13}{##1}}}
\expandafter\def\csname PYGdefault@tok@dl\endcsname{\def\PYGdefault@tc##1{\textcolor[rgb]{0.73,0.13,0.13}{##1}}}
\expandafter\def\csname PYGdefault@tok@s2\endcsname{\def\PYGdefault@tc##1{\textcolor[rgb]{0.73,0.13,0.13}{##1}}}
\expandafter\def\csname PYGdefault@tok@sh\endcsname{\def\PYGdefault@tc##1{\textcolor[rgb]{0.73,0.13,0.13}{##1}}}
\expandafter\def\csname PYGdefault@tok@s1\endcsname{\def\PYGdefault@tc##1{\textcolor[rgb]{0.73,0.13,0.13}{##1}}}
\expandafter\def\csname PYGdefault@tok@mb\endcsname{\def\PYGdefault@tc##1{\textcolor[rgb]{0.40,0.40,0.40}{##1}}}
\expandafter\def\csname PYGdefault@tok@mf\endcsname{\def\PYGdefault@tc##1{\textcolor[rgb]{0.40,0.40,0.40}{##1}}}
\expandafter\def\csname PYGdefault@tok@mh\endcsname{\def\PYGdefault@tc##1{\textcolor[rgb]{0.40,0.40,0.40}{##1}}}
\expandafter\def\csname PYGdefault@tok@mi\endcsname{\def\PYGdefault@tc##1{\textcolor[rgb]{0.40,0.40,0.40}{##1}}}
\expandafter\def\csname PYGdefault@tok@il\endcsname{\def\PYGdefault@tc##1{\textcolor[rgb]{0.40,0.40,0.40}{##1}}}
\expandafter\def\csname PYGdefault@tok@mo\endcsname{\def\PYGdefault@tc##1{\textcolor[rgb]{0.40,0.40,0.40}{##1}}}
\expandafter\def\csname PYGdefault@tok@ch\endcsname{\let\PYGdefault@it=\textit\def\PYGdefault@tc##1{\textcolor[rgb]{0.25,0.50,0.50}{##1}}}
\expandafter\def\csname PYGdefault@tok@cm\endcsname{\let\PYGdefault@it=\textit\def\PYGdefault@tc##1{\textcolor[rgb]{0.25,0.50,0.50}{##1}}}
\expandafter\def\csname PYGdefault@tok@cpf\endcsname{\let\PYGdefault@it=\textit\def\PYGdefault@tc##1{\textcolor[rgb]{0.25,0.50,0.50}{##1}}}
\expandafter\def\csname PYGdefault@tok@c1\endcsname{\let\PYGdefault@it=\textit\def\PYGdefault@tc##1{\textcolor[rgb]{0.25,0.50,0.50}{##1}}}
\expandafter\def\csname PYGdefault@tok@cs\endcsname{\let\PYGdefault@it=\textit\def\PYGdefault@tc##1{\textcolor[rgb]{0.25,0.50,0.50}{##1}}}

\makeatother

\makeatletter
\def\PYG@reset{\let\PYG@it=\relax \let\PYG@bf=\relax%
    \let\PYG@ul=\relax \let\PYG@tc=\relax%
    \let\PYG@bc=\relax \let\PYG@ff=\relax}
\def\PYG@tok#1{\csname PYG@tok@#1\endcsname}
\def\PYG@toks#1+{\ifx\relax#1\empty\else%
    \PYG@tok{#1}\expandafter\PYG@toks\fi}
\def\PYG@do#1{\PYG@bc{\PYG@tc{\PYG@ul{%
    \PYG@it{\PYG@bf{\PYG@ff{#1}}}}}}}
\def\PYG#1#2{\PYG@reset\PYG@toks#1+\relax+\PYG@do{#2}}

\expandafter\def\csname PYG@tok@w\endcsname{\def\PYG@tc##1{\textcolor[rgb]{0.73,0.73,0.73}{##1}}}
\expandafter\def\csname PYG@tok@c\endcsname{\let\PYG@it=\textit\def\PYG@tc##1{\textcolor[rgb]{0.25,0.50,0.50}{##1}}}
\expandafter\def\csname PYG@tok@cp\endcsname{\def\PYG@tc##1{\textcolor[rgb]{0.74,0.48,0.00}{##1}}}
\expandafter\def\csname PYG@tok@k\endcsname{\let\PYG@bf=\textbf\def\PYG@tc##1{\textcolor[rgb]{0.00,0.50,0.00}{##1}}}
\expandafter\def\csname PYG@tok@kp\endcsname{\def\PYG@tc##1{\textcolor[rgb]{0.00,0.50,0.00}{##1}}}
\expandafter\def\csname PYG@tok@kt\endcsname{\def\PYG@tc##1{\textcolor[rgb]{0.69,0.00,0.25}{##1}}}
\expandafter\def\csname PYG@tok@o\endcsname{\def\PYG@tc##1{\textcolor[rgb]{0.40,0.40,0.40}{##1}}}
\expandafter\def\csname PYG@tok@ow\endcsname{\let\PYG@bf=\textbf\def\PYG@tc##1{\textcolor[rgb]{0.67,0.13,1.00}{##1}}}
\expandafter\def\csname PYG@tok@nb\endcsname{\def\PYG@tc##1{\textcolor[rgb]{0.00,0.50,0.00}{##1}}}
\expandafter\def\csname PYG@tok@nf\endcsname{\def\PYG@tc##1{\textcolor[rgb]{0.00,0.00,1.00}{##1}}}
\expandafter\def\csname PYG@tok@nc\endcsname{\let\PYG@bf=\textbf\def\PYG@tc##1{\textcolor[rgb]{0.00,0.00,1.00}{##1}}}
\expandafter\def\csname PYG@tok@nn\endcsname{\let\PYG@bf=\textbf\def\PYG@tc##1{\textcolor[rgb]{0.00,0.00,1.00}{##1}}}
\expandafter\def\csname PYG@tok@ne\endcsname{\let\PYG@bf=\textbf\def\PYG@tc##1{\textcolor[rgb]{0.82,0.25,0.23}{##1}}}
\expandafter\def\csname PYG@tok@nv\endcsname{\def\PYG@tc##1{\textcolor[rgb]{0.10,0.09,0.49}{##1}}}
\expandafter\def\csname PYG@tok@no\endcsname{\def\PYG@tc##1{\textcolor[rgb]{0.53,0.00,0.00}{##1}}}
\expandafter\def\csname PYG@tok@nl\endcsname{\def\PYG@tc##1{\textcolor[rgb]{0.63,0.63,0.00}{##1}}}
\expandafter\def\csname PYG@tok@ni\endcsname{\let\PYG@bf=\textbf\def\PYG@tc##1{\textcolor[rgb]{0.60,0.60,0.60}{##1}}}
\expandafter\def\csname PYG@tok@na\endcsname{\def\PYG@tc##1{\textcolor[rgb]{0.49,0.56,0.16}{##1}}}
\expandafter\def\csname PYG@tok@nt\endcsname{\let\PYG@bf=\textbf\def\PYG@tc##1{\textcolor[rgb]{0.00,0.50,0.00}{##1}}}
\expandafter\def\csname PYG@tok@nd\endcsname{\def\PYG@tc##1{\textcolor[rgb]{0.67,0.13,1.00}{##1}}}
\expandafter\def\csname PYG@tok@s\endcsname{\def\PYG@tc##1{\textcolor[rgb]{0.73,0.13,0.13}{##1}}}
\expandafter\def\csname PYG@tok@sd\endcsname{\let\PYG@it=\textit\def\PYG@tc##1{\textcolor[rgb]{0.73,0.13,0.13}{##1}}}
\expandafter\def\csname PYG@tok@si\endcsname{\let\PYG@bf=\textbf\def\PYG@tc##1{\textcolor[rgb]{0.73,0.40,0.53}{##1}}}
\expandafter\def\csname PYG@tok@se\endcsname{\let\PYG@bf=\textbf\def\PYG@tc##1{\textcolor[rgb]{0.73,0.40,0.13}{##1}}}
\expandafter\def\csname PYG@tok@sr\endcsname{\def\PYG@tc##1{\textcolor[rgb]{0.73,0.40,0.53}{##1}}}
\expandafter\def\csname PYG@tok@ss\endcsname{\def\PYG@tc##1{\textcolor[rgb]{0.10,0.09,0.49}{##1}}}
\expandafter\def\csname PYG@tok@sx\endcsname{\def\PYG@tc##1{\textcolor[rgb]{0.00,0.50,0.00}{##1}}}
\expandafter\def\csname PYG@tok@m\endcsname{\def\PYG@tc##1{\textcolor[rgb]{0.40,0.40,0.40}{##1}}}
\expandafter\def\csname PYG@tok@gh\endcsname{\let\PYG@bf=\textbf\def\PYG@tc##1{\textcolor[rgb]{0.00,0.00,0.50}{##1}}}
\expandafter\def\csname PYG@tok@gu\endcsname{\let\PYG@bf=\textbf\def\PYG@tc##1{\textcolor[rgb]{0.50,0.00,0.50}{##1}}}
\expandafter\def\csname PYG@tok@gd\endcsname{\def\PYG@tc##1{\textcolor[rgb]{0.63,0.00,0.00}{##1}}}
\expandafter\def\csname PYG@tok@gi\endcsname{\def\PYG@tc##1{\textcolor[rgb]{0.00,0.63,0.00}{##1}}}
\expandafter\def\csname PYG@tok@gr\endcsname{\def\PYG@tc##1{\textcolor[rgb]{1.00,0.00,0.00}{##1}}}
\expandafter\def\csname PYG@tok@ge\endcsname{\let\PYG@it=\textit}
\expandafter\def\csname PYG@tok@gs\endcsname{\let\PYG@bf=\textbf}
\expandafter\def\csname PYG@tok@gp\endcsname{\let\PYG@bf=\textbf\def\PYG@tc##1{\textcolor[rgb]{0.00,0.00,0.50}{##1}}}
\expandafter\def\csname PYG@tok@go\endcsname{\def\PYG@tc##1{\textcolor[rgb]{0.53,0.53,0.53}{##1}}}
\expandafter\def\csname PYG@tok@gt\endcsname{\def\PYG@tc##1{\textcolor[rgb]{0.00,0.27,0.87}{##1}}}
\expandafter\def\csname PYG@tok@err\endcsname{\def\PYG@bc##1{\setlength{\fboxsep}{0pt}\fcolorbox[rgb]{1.00,0.00,0.00}{1,1,1}{\strut ##1}}}
\expandafter\def\csname PYG@tok@kc\endcsname{\let\PYG@bf=\textbf\def\PYG@tc##1{\textcolor[rgb]{0.00,0.50,0.00}{##1}}}
\expandafter\def\csname PYG@tok@kd\endcsname{\let\PYG@bf=\textbf\def\PYG@tc##1{\textcolor[rgb]{0.00,0.50,0.00}{##1}}}
\expandafter\def\csname PYG@tok@kn\endcsname{\let\PYG@bf=\textbf\def\PYG@tc##1{\textcolor[rgb]{0.00,0.50,0.00}{##1}}}
\expandafter\def\csname PYG@tok@kr\endcsname{\let\PYG@bf=\textbf\def\PYG@tc##1{\textcolor[rgb]{0.00,0.50,0.00}{##1}}}
\expandafter\def\csname PYG@tok@bp\endcsname{\def\PYG@tc##1{\textcolor[rgb]{0.00,0.50,0.00}{##1}}}
\expandafter\def\csname PYG@tok@fm\endcsname{\def\PYG@tc##1{\textcolor[rgb]{0.00,0.00,1.00}{##1}}}
\expandafter\def\csname PYG@tok@vc\endcsname{\def\PYG@tc##1{\textcolor[rgb]{0.10,0.09,0.49}{##1}}}
\expandafter\def\csname PYG@tok@vg\endcsname{\def\PYG@tc##1{\textcolor[rgb]{0.10,0.09,0.49}{##1}}}
\expandafter\def\csname PYG@tok@vi\endcsname{\def\PYG@tc##1{\textcolor[rgb]{0.10,0.09,0.49}{##1}}}
\expandafter\def\csname PYG@tok@vm\endcsname{\def\PYG@tc##1{\textcolor[rgb]{0.10,0.09,0.49}{##1}}}
\expandafter\def\csname PYG@tok@sa\endcsname{\def\PYG@tc##1{\textcolor[rgb]{0.73,0.13,0.13}{##1}}}
\expandafter\def\csname PYG@tok@sb\endcsname{\def\PYG@tc##1{\textcolor[rgb]{0.73,0.13,0.13}{##1}}}
\expandafter\def\csname PYG@tok@sc\endcsname{\def\PYG@tc##1{\textcolor[rgb]{0.73,0.13,0.13}{##1}}}
\expandafter\def\csname PYG@tok@dl\endcsname{\def\PYG@tc##1{\textcolor[rgb]{0.73,0.13,0.13}{##1}}}
\expandafter\def\csname PYG@tok@s2\endcsname{\def\PYG@tc##1{\textcolor[rgb]{0.73,0.13,0.13}{##1}}}
\expandafter\def\csname PYG@tok@sh\endcsname{\def\PYG@tc##1{\textcolor[rgb]{0.73,0.13,0.13}{##1}}}
\expandafter\def\csname PYG@tok@s1\endcsname{\def\PYG@tc##1{\textcolor[rgb]{0.73,0.13,0.13}{##1}}}
\expandafter\def\csname PYG@tok@mb\endcsname{\def\PYG@tc##1{\textcolor[rgb]{0.40,0.40,0.40}{##1}}}
\expandafter\def\csname PYG@tok@mf\endcsname{\def\PYG@tc##1{\textcolor[rgb]{0.40,0.40,0.40}{##1}}}
\expandafter\def\csname PYG@tok@mh\endcsname{\def\PYG@tc##1{\textcolor[rgb]{0.40,0.40,0.40}{##1}}}
\expandafter\def\csname PYG@tok@mi\endcsname{\def\PYG@tc##1{\textcolor[rgb]{0.40,0.40,0.40}{##1}}}
\expandafter\def\csname PYG@tok@il\endcsname{\def\PYG@tc##1{\textcolor[rgb]{0.40,0.40,0.40}{##1}}}
\expandafter\def\csname PYG@tok@mo\endcsname{\def\PYG@tc##1{\textcolor[rgb]{0.40,0.40,0.40}{##1}}}
\expandafter\def\csname PYG@tok@ch\endcsname{\let\PYG@it=\textit\def\PYG@tc##1{\textcolor[rgb]{0.25,0.50,0.50}{##1}}}
\expandafter\def\csname PYG@tok@cm\endcsname{\let\PYG@it=\textit\def\PYG@tc##1{\textcolor[rgb]{0.25,0.50,0.50}{##1}}}
\expandafter\def\csname PYG@tok@cpf\endcsname{\let\PYG@it=\textit\def\PYG@tc##1{\textcolor[rgb]{0.25,0.50,0.50}{##1}}}
\expandafter\def\csname PYG@tok@c1\endcsname{\let\PYG@it=\textit\def\PYG@tc##1{\textcolor[rgb]{0.25,0.50,0.50}{##1}}}
\expandafter\def\csname PYG@tok@cs\endcsname{\let\PYG@it=\textit\def\PYG@tc##1{\textcolor[rgb]{0.25,0.50,0.50}{##1}}}

\def\PYGZbs{\char`\\}
\def\PYGZus{\char`\_}
\def\PYGZob{\char`\{}
\def\PYGZcb{\char`\}}

\def\PYGZsh{\char`\#}

\def\PYGZhy{\char`\-}

\def\PYGZdq{\char`\"}

\makeatother

\jmlrheading{1}{2020}{1-48}{4/00}{10/00}{meila00a}{Théophile Champion and Marek Grze\'s and Howard Bowman}

\ShortHeadings{Multi-Modal and Multi-Factor BTAI.}{Champion et al.}
\firstpageno{1}

\begin{document}

\title{Multi-Modal and Multi-Factor Branching Time Active Inference $(BTAI_{3MF}$).}

\author{\name Théophile Champion \email tmac3@kent.ac.uk \\
       \addr University of Kent, School of Computing\\
       Canterbury CT2 7NZ, United Kingdom
       \AND
       \name Marek Grze\'s \email m.grzes@kent.ac.uk \\
       \addr University of Kent, School of Computing\\
       Canterbury CT2 7NZ, United Kingdom
       \AND
       \name Howard Bowman \email H.Bowman@kent.ac.uk \\
       \addr University of Birmingham, School of Psychology,\\
       Birmingham B15 2TT, United Kingdom\\
       University of Kent, School of Computing\\
       Canterbury CT2 7NZ, United Kingdom
       }
       
\editor{\textbf{TO BE FILLED}} 

\maketitle

\begin{abstract}
Active inference is a state-of-the-art framework for modelling the brain that explains a wide range of mechanisms such as habit formation, dopaminergic discharge and curiosity. Recently, two versions of branching time active inference (BTAI) based on Monte-Carlo tree search have been developed to handle the exponential (space and time) complexity class that occurs when computing the prior over all possible policies up to the time horizon. However, those two versions of BTAI still suffer from an exponential complexity class w.r.t the number of observed and latent variables being modelled. In the present paper, we resolve this limitation by first allowing the modelling of several observations, each of them having its own likelihood mapping. Similarly, we allow each latent state to have its own transition mapping. The inference algorithm then exploits the factorisation of the likelihood and transition mappings to accelerate the computation of the posterior. Those two optimisations were tested on the dSprites environment in which the metadata of the dSprites dataset was used as input to the model instead of the dSprites images. On this task, $BTAI_{VMP}$ \citep{AITS_THEORY, AITS_PRACTICE} was able to solve 96.9\% of the task in 5.1 seconds, and $BTAI_{BF}$ \citep{BTAI_BF} was able to solve 98.6\% of the task in 17.5 seconds. Our new approach ($BTAI_{3MF}$) outperformed both of its predecessors by solving the task completly (100\%) in only 2.559 seconds. Finally, $BTAI_{3MF}$ has been implemented in a flexible and easy to use (python) package, and we developed a graphical user interface to enable the inspection of the model's beliefs, planning process and behaviour.
\end{abstract}


\section{Introduction}

Active inference extends the free energy principle to generative models with actions \citep{FRISTON2016862,AI_TUTO,AI_VMP} and can be regarded as a form of planning as inference \citep{PAI}. This framework has successfully explained a wide range of neuro-cognitive phenomena, such as habit formation \citep{FRISTON2016862}, Bayesian surprise \citep{bayes_surprise}, curiosity \citep{curiosity}, and dopaminergic discharges \citep{dopamine}. It has also been applied to a variety of tasks, such as animal navigation \citep{DeepAIwithMCMC}, robotic control \citep{pezzato2020active,sancaktar2020endtoend}, the mountain car problem \citep{catal2020learning}, the  game DOOM \citep{CULLEN2018809} and the cart pole problem \citep{cart_pole}.

However, active inference suffers from an exponential (space and time) complexity class that occurs when computing the prior over all possible policies up to the time horizon. Recently, two versions of branching time active inference (BTAI) based on Monte-Carlo tree search \citep{MCTS} have been developed to handle this exponential growth. In the original formulation of the framework \citep{AITS_THEORY, AITS_PRACTICE}, inference was performed using the variational message passing (VMP) algorithm \citep{VMP_TUTO, AI_VMP}. In a follow up paper, VMP was then replaced by a Bayesian filtering \citep{BAYESIAN_FILTERING} scheme leading to a faster inference process \citep{BTAI_BF}.

In this paper, we develop an extension of Branching Time Active Inference (BTAI), to allow modelling of several modalities as well as several latent states. Indeed, even if the Bayesian filtering version of Branching Time Active Inference ($BTAI_{BF}$) is fast, its modelling capacity is limited to one observation and one hidden state. Consequently, if one wanted to model $n$ latent states $S_t^1, \hdots, S_t^n$, then those $n$ latent states would have to be encoded into one latent state $X$ representing all possible configurations of the $n$ latent states $S_t^1, \hdots, S_t^n$. Unfortunatly, the total number of configurations is given by:
\begin{align*}
\nb{X} = \prod_{i=1}^n \nb{S_t^i} \geq 2^n,
\end{align*}
where $\nb{X}$ is the number of possible values taken by $X$, and similarly $\nb{S_t^i}$ is the number of possible values taken by $S_t^i$. The above inequality is obtained by realizing that $\nb{S_t^i} \geq 2$, and is problematic in practice because $\nb{X}$ is growing exponentially with the number of latent states $n$ being modelled. Also, note that in practice this exponential growth may be way worse than $2^n$. For example, 
if one were to model the five modalities of the dSprites environment (c.f. Section \ref{ssec:dsprites}), the total number of configurations would be:
$$\nb{S^y_t} \times \nb{S^{x}_t} \times \nb{S^{scale}_t} \times \nb{S^{orientation}_t} \times \nb{S^{scale}_t} = 33 \times 32 \times 3 \times 40 \times 6 = 760,320 \gg 2^5 = 32.$$
A similar exponential explosion also appears when trying to model several modalities $O_t^1, \hdots, O_t^m$ using a single one $Y$, i.e.
\begin{align*}
\nb{Y} = \prod_{i=1}^m \nb{O_t^i} \geq 2^m,
\end{align*}
where $\nb{Y}$ is the number of possible values taken by $Y$, and similarly $\nb{O_t^i}$ is the number of possible values taken by $O_t^i$. Note, throughout this paper, we will use the term \textit{states} to refer to the latent states of the model at a specify time step, e.g., $S_t^1, \hdots, S_t^n$ for time step $t$. Additionally, we will use the terms \textit{state configurations} or \textit{values} to refer to particular values taken by the latent variables.

The present paper aims to remove those two exponential growths, by allowing the modelling of several observations and latent states, while providing an easy to use framework based on a high-level notation, which allows the user to create models by simply declaring the variables it contains, and the dependencies between those variables. Then, the framework performs the inference process automatically. Appendix A shows an example of how to implement a custom $BTAI_{3MF}$ agent using our framework. In section \ref{sec:btai_3mf}, we describe the theory underlying our approach. Importantly, $BTAI_{3MF}$ takes advantage of the generative model struture to perform inference efficiently using a mixture of belief propagation \citep{BP_and_DC, believe,belief_propagation} and forward predictions as will be explained in Section \ref{ssec:IP_algorithm}. The name $BTAI_{3MF}$ is an abbreviation for  $BTAI_{MMMF}$ that stands for: Multi-Modal and Multi-Factor Branching Time Active Inference. Next, in Section \ref{ssec:efe}, we provide the definition of the expected free energy in the context of our new approach, and in Section \ref{ssec:planning}, we describe the planning algorithm used to expand the generative model dynamically. Then, in Section \ref{sec:results}, we compare $BTAI_{3MF}$ to $BTAI_{VMP}$ and $BTAI_{BF}$, and demonstrate empirically that $BTAI_{3MF}$ outperformed both $BTAI_{VMP}$ and $BTAI_{BF}$ on the dSprites environment, which requires the modelling of many latent states and modalities. Finally, Section \ref{sec:conclusion} concludes this paper by summarizing our approach and results.

\section{Theory of $BTAI_{3MF}$} \label{sec:btai_3mf}

In this section, we introduce the mathematical foundation of $BTAI_{3MF}$. To simplify the graphical representation of our generative model, we first introduce a notion of ``temporal slice". Then, we build on this idea to describe the generative model of $BTAI_{3MF}$. Next, we explain how belief updates are performed using a mixture of belief propagation and forward predictions. Afterwards, we provide the definition of the expected free energy for this new generative model. Finally, we describe the planning algorithm used to dynamically expand the generative model, and the action selection process.

\subsection{Temporal slice} \label{ssec:temporal_slice}

A temporal slice $TS_J = \{O_J^1, \hdots, O_J^{\nb{O}}, S_J^1, \hdots, S_J^{\nb{S}}\}$ is a set of random variables indexed by a sequence of actions $J$. Each random variable of the temporal slice represents either an observation $O_J^o$ or a latent state $S_J^s$. The index of the temporal slice correponds to the sequence of actions that lead to this temporal slice. By definition, if $J$ is an empty sequence, i.e., $J = \emptyset$, then $TS_J$ is the temporal slice of the present time step $t$, also denoted $TS_t$. Within a temporal slice $TS_J$, an observation $O_J^o$ depends on a number of latent states $\rho_J^o \subseteq \{S_J^s \mid s = 1, \hdots, \nb{S}\}$, such that $P(O_J^o|\rho_J^o)$ is a factor in the generative model. Given an action $\bm{a}$ and a sequence of actions $J$, we let $I = J{::}\bm{a}$ be the sequence of actions obtained by appending the action $\bm{a}$ at the end of the sequence of actions $J$. If $I = J{::}\bm{a}$, then the temporal slice $TS_J$ can be the parent of $TS_I$. This means that a latent state $S^s_I$ in $TS_I$ can depend on the latent states $\rho_I^s \subseteq \{S_J^s \mid s = 1, \hdots, \nb{S}\}$ in $TS_J$, such that $P(S_I^s|\rho_I^s)$ is a factor in the generative model. The concept of temporal slice is illustrated in Figure \ref{fig:temporal_slice}, and Figure \ref{fig:temporal_slice_compact} depicts a more compact representation of the content of Figure \ref{fig:temporal_slice}.

\begin{figure}[H]
	\begin{center}
	\begin{tikzpicture}[square/.style={regular polygon,regular polygon sides=4}]
		\node (TS_I) at (0,-0.25) [rectangle, draw, minimum width=3cm, minimum height=4.7cm, very thick] {};
		\node[below=of TS_I,yshift=0.5cm] (TS_I_label) {$TS_t$};
		\draw[black, very thick] (TS_I) -- (TS_I_label);
		
        \node[latent] (SI) at (0,0.5) {$S_t^s$};
		\plate[inner xsep=1cm, inner ysep=0.5cm, yshift=0.5cm] {plate_SI} {(SI)} {};
        \node[] (SI_label) at (0,1.5) {$s = 1, \hdots, \nb{S}$};

        \node[obs] (OI) at (0,-1) {$O_t^o$};
		\plate[inner xsep=1cm, inner ysep=0.5cm, yshift=-0.3cm] {plate_OI} {(OI)} {};
        \node[] (OI_label) at (0,-2) {$o = 1, \hdots, \nb{O}$};

		\draw [densely dashed,-latex] (SI) -- (OI);

		\node (TS_J) at (4.05,-0.25) [right=of TS_I, rectangle, draw, minimum width=3cm, minimum height=4.7cm, very thick] {};
		\node[below=of TS_J,yshift=0.5cm] (TS_J_label) {$TS_I$};
		\draw[black, very thick] (TS_J) -- (TS_J_label);
		
        \node[latent] (SJ) at (4.05,0.5) {$S_I^s$};
		\plate[inner xsep=1cm, inner ysep=0.5cm, yshift=0.5cm] {plate_SJ} {(SJ)} {};
        \node[] (SJ_label) at (4.05,1.5) {$s = 1, \hdots, \nb{S}$};

        \node[latent] (OJ) at (4.05,-1) {$O_I^o$};
		\plate[inner xsep=1cm, inner ysep=0.5cm, yshift=-0.3cm] {plate_OJ} {(OJ)} {};
        \node[] (OJ_label) at (4.05,-2) {$o = 1, \hdots, \nb{O}$};

		\draw [densely dashed,-latex] (SJ) -- (OJ);

		\draw [densely dashed,-latex] (SI) -- (SJ);
    \end{tikzpicture}
 	\end{center}
\vspace{-0.25cm}
    \caption{
This figure illustrates two temporal slices $TS_t$ and $TS_I$, which are depicted by rectangles with thick border. Within each temporal slice, plate notation is used to generate $\nb{S}$ latent states and $\nb{O}$ observations. The dashed lines that connect two random variables from two different plates are new to this paper, and represent an arbitrary connectivity between the two sets of random variables generated by the plates. For example, the dashed line from $S_t^s$ to $O_t^o$, means that for each observation $O_t^o$, the parents of $O_t^o$ denoted $\rho_t^o$ is a subset of $\{S_t^s \mid s = 1, \hdots, \nb{S}\}$, i.e., the generative model contains the factor $P(O_t^o | \rho_t^o)$ where $\rho_t^o \subseteq \{S_t^s \mid s = 1, \hdots, \nb{S}\}$.
}
    \label{fig:temporal_slice}
\end{figure}
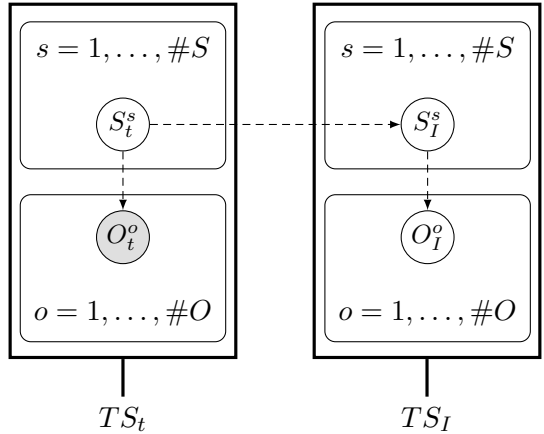

\begin{figure}[H]
	\begin{center}
	\begin{tikzpicture}[square/.style={regular polygon,regular polygon sides=4}]
		\node (TS_I) at (0,-0.25) [rectangle, fill=gray!20, draw, minimum width=1cm, minimum height=1cm, very thick] {$TS_t$};
		
		\node (TS_J) at (4.05,-0.25) [right=of TS_I, rectangle, draw, minimum width=1cm, minimum height=1cm, very thick] {$TS_I$};

		\draw [densely dashed,-latex] (TS_I) -- (TS_J);
    \end{tikzpicture}
 	\end{center}
\vspace{-0.25cm}
    \caption{
This figure illustrates the two temporal slices $TS_t$ and $TS_I$ from Figure \ref{fig:temporal_slice} in a more compact fashion. Since $O_t^o$ is an observed variable for all $o \in \{1, \hdots, \nb{O}\}$, the square representing $TS_t$ has a gray background. In contrast, the square representing $TS_I$ has a white background because $O_I^o$ is a latent variable for all $o \in \{1, \hdots, \nb{O}\}$.
}
    \label{fig:temporal_slice_compact}
\end{figure}
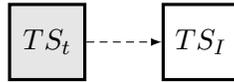

\subsection{Generative model}

\begin{figure}[H]
	\begin{center}
	\begin{tikzpicture}[square/.style={regular polygon,regular polygon sides=4}]
		\node (TS_I) at (0,-0.25) [rectangle, fill=gray!20, draw, minimum width=1.5cm, minimum height=1.5cm, very thick] {$TS_t$};
		
		\node (TS_1) at (4.05,-0.25) [below=of TS_I, rectangle, draw, minimum width=1.5cm, minimum height=1.5cm, very thick, xshift=-2cm] {$TS_{(1)}$};
		\node (TS_2) at (4.05,-0.25) [below=of TS_I, rectangle, draw, minimum width=1.5cm, minimum height=1.5cm, very thick, xshift=2cm] {$TS_{(2)}$};

		\node (TS_11) at (4.05,-0.25) [below=of TS_1, rectangle, draw, minimum width=1.5cm, minimum height=1.5cm, very thick, xshift=-1cm] {$TS_{(11)}$};
		\node (TS_12) at (4.05,-0.25) [below=of TS_1, rectangle, draw, minimum width=1.5cm, minimum height=1.5cm, very thick, xshift=1cm] {$TS_{(12)}$};

		\node (TS_21) at (4.05,-0.25) [below=of TS_2, rectangle, draw=gray!50, minimum width=1.5cm, minimum height=1.5cm, very thick, xshift=-1cm] {${\color{gray!50}TS_{(21)}}$};
		\node (TS_22) at (4.05,-0.25) [below=of TS_2, rectangle, draw=gray!50, minimum width=1.5cm, minimum height=1.5cm, very thick, xshift=1cm] {${\color{gray!50}TS_{(22)}}$};

		\draw [densely dashed,-latex] (TS_I) -- (TS_1);
		\draw [densely dashed,-latex] (TS_I) -- (TS_2);
		\draw [densely dashed,-latex] (TS_1) -- (TS_11);
		\draw [densely dashed,-latex] (TS_1) -- (TS_12);
		\draw [densely dashed,-latex,draw=gray!50] (TS_2) -- (TS_21);
		\draw [densely dashed,-latex,draw=gray!50] (TS_2) -- (TS_22);
		
		\node (I) at (5,-0.25) {$\mathbb{I} = \Big\{(1), (2), (11), (12)\Big\}$};
		
    \end{tikzpicture}
 	\end{center}
\vspace{-0.25cm}
    \caption{
This figure illustrates the full generative model of $BTAI_{3MF}$. The temporal slices depited in light gray correspond to temporal slices that have not yet been explored by the planning algorithm, c.f., Section \ref{ssec:planning}. The numbers between parentheses correspond to the sequence of actions performed to reach the temporal slice.
}
    \label{fig:generative_model}
\end{figure}
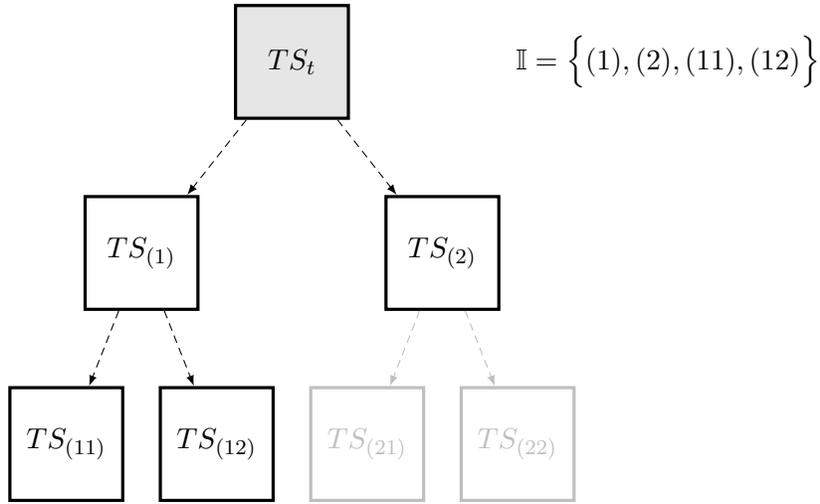

In this section, we build upon the notion of temporal slice to describe the full generative model. Intuitively, the probability of the entire generative model is the product of the probability of each temporal slice within the model. This includes the current temporal slice $TS_t$ and the future temporal slices $TS_I$ for all $I \in \mathbb{I}$, where $\mathbb{I}$ is the set of all multi-indices expanded during the tree search (c.f., Section \ref{ssec:planning}). Within each temporal slice, there are $\nb{O}$ observations and $\nb{S}$ latent states. Each observation depends on a subset of the latent states. Moreover, each latent state depends on a subset of the latent states of the parent temporal slice. Note, the current temporal slice $TS_t$ does not have any parents, therefore its latent state does not depend on anything. In other words, the model makes the Markov assumption, i.e., each state only depends on the states at the previous time step. More formally, the generative model is defined as:
\begin{align*}
P(O_t,S_t,O_\mathbb{I},S_\mathbb{I}) &= P(TS_t) \prod_{I\in\mathbb{I}} P(TS_I)\\
&= \underbrace{\prod_{o=1}^{\nb{O}} P(O_t^o|\rho_t^o)\prod_{s=1}^{\nb{S}} P(S_t^s)}_{\text{current temporal slice }TS_t} \prod_{I\in\mathbb{I}} \Bigg[ \underbrace{\prod_{o=1}^{\nb{O}} P(O_I^o|\rho_I^o)\prod_{s=1}^{\nb{S}} P(S_I^s|\rho_I^s)}_{\text{future temporal slice }TS_I} \Bigg]
\end{align*}
where $t$ is the current time step, $\rho_\tau^x$ is the set of parents of $X^x_\tau$, $O_t = \{O_t^o \mid o = 1, \hdots, \nb{O}\}$ is the set of all observations at time $t$, $O_I = \{O_I^o \mid o = 1, \hdots, \nb{O}\}$ is the set of all future observations that would be observed after performing the sequence of actions $I$, $O_\mathbb{I} = \cup_{I \in \mathbb{I}} O_I$ is the set of all future observations contained in the temporal slices expanded during the tree search (c.f., Section \ref{ssec:planning}), $S_t = \{S_t^s \mid s = 1, \hdots, \nb{S}\}$ is the set of all latent states at time $t$, $S_I = \{S_I^s \mid s = 1, \hdots, \nb{S}\}$ is the set of random variables describing the future latent states after performing the sequence of actions $I$, $S_\mathbb{I} = \cup_{I \in \mathbb{I}} S_I$ is the set of latent variables representing all future states contained in the temporal slices expanded during the tree search (c.f., Section \ref{ssec:planning}). Importantly, the above generative model has to satisfy:
\begin{itemize}
\item $\forall I \in \mathbb{I}, \forall o \in \{1, \hdots, \nb{O}\}, \rho_I^o \subseteq S_I$;
\item $\forall I{::}\bm{a} \in \mathbb{I}, \forall s \in \{1, \hdots, \nb{S}\}, \rho_{I{::}\bm{a}}^s \subseteq S_I$, also, if $I = \emptyset$ then by definition $S_I \delequal S_t$.
\end{itemize}
\noindent Additionally, we define the factors of the generative model as:
\begin{align*}
P(O_t^o|\rho_t^o) = \text{Cat}(\bm{A}^o), & \qquad P(S_t^s) = \text{Cat}(\bm{D}^s_t),\\
P(O_I^o|\rho_I^o) = \text{Cat}(\bm{A}^o), & \qquad P(S_I^s|\rho_I^s) = \text{Cat}(\bm{B}^s_I),
\end{align*}
where $\bm{A}^o$ is the tensor modelling the likelihood mapping of the $o$-th observation, $\bm{D}^s_t$ is the vector modelling the prior over the $s$-th latent state at time $t$ (see below for details), $\bm{B}^s$ is the tensor modelling the transition mapping of the $s$-th latent state under each possible action, $\bm{B}^s_I$ is the tensor modelling the transition mapping of the $s$-th latent state under the last action $I_\text{last}$ of the sequence $I$, i.e., $\bm{B}^s_I = \bm{B}^s(\, \bigcdot \,, \hdots, \bigcdot\, , I_\text{last})$. Also, note that at the beginning of a trial, i.e., when $t=0$, $\bm{D}^s_t$ is a vector that encodes the modeller's understanding of the task. Afterwards, when $t > 0$, $\bm{D}^s_t$ is a vector containing the parameters of the posterior over hidden states according to the observations made and actions taken so far, i.e., $P(S_t^s) \delequal P(S_t^s|O_{0:t-1}, A_{0:t-1}) = \text{Cat}(\bm{D}^s_t)$ for all $s \in \{1, \hdots, \nb{S}\}$. Finally, Figure \ref{fig:generative_model} illustrates the full generative model using the notion of temporal slices.

\subsection{Belief updates: the inference and prediction (IP) algorithm} \label{ssec:IP_algorithm}

The IP algorithm is composed of two steps, i.e., the inference step (or I-step) and the prediction step (or P-step). The goal of the I-step is to compute the posterior beliefs over all the latent variables at time $t$. In other words, the goal of the I-step is to compute: $P(S_t^s|O_t), \forall s \in \{1, \hdots, \nb{S}\}$. The P-step takes as inputs the posterior beliefs over all the latent variables corresponding to the states of the system after performing a sequence of actions $I$, and an action $\bm{a}$ to be performed next. The goal of the P-step is to compute the posterior beliefs over all the latent variables corresponding to the future states and observations after performing the sequence of actions $I{::}\bm{a}$, where $I{::}\bm{a}$ is the sequence of actions obtained by adding the action $\bm{a}$ at the end of the sequence of actions $I$. In other words, given $P(S_I^s|O_t), \forall s \in \{1, \hdots, \nb{S}\}$ and an action $\bm{a}$, the goal of the P-step is to compute: $P(S_{I{::}\bm{a}}^s|O_t), \forall s \in \{1, \hdots, \nb{S}\}$ and $P(O_{I{::}\bm{a}}^o|O_t), \forall o \in \{1, \hdots, \nb{O}\}$. Note that by definition, we let $P(S_I^m|O_t) \delequal P(S_t^m|O_t)$ if $I = \emptyset$. To derive the inference and prediction steps, the following sections make use of the sum-rule, product-rule, and d-separation criterion (c.f., Appendix C for details about those properties).

\subsubsection{Inference step} \label{ssec:i_step}

As just stated, the goal of the I-step is to compute $P(S_t^m|O_t), \forall m \in \{1, \hdots, \nb{S}\}$. First, we re-write the posterior computation to fit the kind of problem that belief propagation --- also known as the sum-product algorithm --- can solve:
\begin{align*}
P(S_t^m|O_t) &\propto P(S^m_t, O_t){\color{white}\sum_{\sim S^m_t}^T} \tag{Bayes theorem}\\
&= \sum_{\sim S^m_t}^{{\color{white}}} P(S_t, O_t) \tag{sum rule}\\
&= \sum_{\sim S^m_t} \prod_{o=1}^{\nb{O}} P(O_t^o|\rho_t^o)\prod_{s=1}^{\nb{S}} P(S_t^s) \tag{product rule \& d-separation}\\
\end{align*}
where $S_t = \{S_t^s \mid s = 1, \hdots, \nb{S}\}$ is the set of all latent states at time $t$, ${\sim}S_t^m = S_t \setminus S_t^m$ is the set of all latent states at time $t$ except $S^m_t$, and the summation is over all possible configurations of ${\sim}S_t^m$, i.e., we are marginalizing out all states, apart from one; thus $P(S_t, O_t)$ has $\nb{S} + \nb{O}$ dimensions, while $P(S^m_t, O_t)$  has $1 + \nb{O}$ dimensions. Since $\rho_t^o \subseteq S_t$, the expression inside the summation is a function $g(S_t)$ that factorizes as follows:
\begin{align*}
g(S_t) &= \prod_{o=1}^{\nb{O}} P(O_t^o|\rho_t^o) \prod_{s=1}^{\nb{S}} P(S_t^s)\\
&\delequal \prod_{i=1}^{N} f_i(X_i),
\end{align*}
where $X_i \subseteq S_t$ for all $i \in \{1, \hdots, \nb{O} + \nb{S}\}$, the number of factors is $N = \nb{O}+\nb{S}$, and:
\begin{align*}
f_i(X_i) \delequal \begin{cases}
	P(O_t^i|\rho_t^i) & \text{if } i \in \{1, \hdots, \nb{O}\}\\
	P(S_t^{i - \nb{O}}) & \text{if } i \in \{\nb{O} + 1, \hdots, \nb{O}+\nb{S}\}\\
\end{cases}.
\end{align*}
Note that, because $O_t^o$ (denoted $O_t^i$ here) are known constants, we do not specify that $g(S_t)$ depends on $O_t^o$. To conclude, by substituting the definition of $g(S_t)$ into the formula of the posterior $P(S_t^m|O_t)$ presented above, we get:
\begin{align*}
P(S_t^m|O_t) &\propto \sum_{\sim S^m_t} g(S_t),
\end{align*}
which means that the posterior $P(S_t^m|O_t)$ can be computed by first marginalizing $g(S_t)$ w.r.t. $S_t^m$, i.e.,
\begin{align*}
g(S_t^m) = \sum_{\sim S^m_t} g(S_t),
\end{align*}
and then normalizing:
\begin{align*}
P(S_t^m|O_t) = \frac{g(S_t^m)}{\sum_{S^m_t} g(S^m_t)}.
\end{align*}
The marginalization of $g(S_t)$ can be performed efficiently using belief propagation \citep{belief_propagation}, which can be understood as a message passing algorithm on a factor graph. The message from a node $x$ to a factor $f$ is given by:
\begin{align*}
m_{x \rightarrow f}(x) = \prod_{h \in n(x) \setminus \{f\}} m_{h \rightarrow x}(x),
\end{align*}
where $n(x)$ are the neighbours of $x$ in the factor graph. Note, in a factor graph the neighbours of a random variable are factors. Moreover, the message from a factor $f$ to a node $x$ is given by:
\begin{align*}
m_{f \rightarrow x}(x) = \sum_{Y} \Bigg( f(X) \prod_{y \in Y} m_{y \rightarrow f}(y)\Bigg),
\end{align*}
where $X = n(f)$ are the neighbours of $f$ in the factor graph, $Y = X \setminus \{x\}$ are all the neighbours of $f$ except $x$, and the summation is over all possible configurations of the variables in $Y$. Note, in a factor graph the neighbours of a factor are random variables. Once all the messages have been computed, the marginalization of $g(S_t)$ w.r.t. $S_t^m$ is given by the product of all the incoming messages of the node $S_t^m$, i.e.,
\begin{align*}
g(S_t^m) = \prod_{f \in n(S_t^m)} m_{f \rightarrow S_t^m}(S_t^m).
\end{align*}

\subsubsection{Prediction step} \label{ssec:p_step}

The P-step is analogous to the prediction step of Bayesian filtering \citep{BAYESIAN_FILTERING}. Given $P(S_{I}^s|O_t)$ for each $s \in \{1, \hdots, \nb{S}\}$ and an action $\bm{a}$, the goal of the P-step is to compute $P(S_{I{::}\bm{a}}^s|O_t)$ for each latent state $s \in \{1, \hdots, \nb{S}\}$ and $P(O_{I{::}\bm{a}}^o|O_t)$ for each future observation $o \in \{1, \hdots, \nb{O}\}$. For the sake of brevity, we let $J \delequal I{::}\bm{a}$. Let's start with the computation of $P(S_{I{::}\bm{a}}^s|O_t)$:
\begin{align*}
P(S_{I{::}\bm{a}}^s|O_t) \delequal P(S_J^s|O_t) &= \sum_{\rho_J^s}^{{\color{white}M}} P(S_J^s, \rho_J^s |O_t) \tag{sum rule}\\
&= \sum_{\rho_J^s}^{{\color{white}M}} P(S_J^s| \rho_J^s, O_t)P(\rho_J^s| O_t) \tag{product rule}\\
&= \sum_{\rho_J^s}^{{\color{white}M}} P(S_J^s| \rho_J^s)P(\rho_J^s| O_t) \tag{d-separation}\\
&\approx \sum_{\rho_J^s} P(S_J^s| \rho_J^s) \prod_{i=1}^{\nb{\rho_J^s}} P(\rho_{J,i}^s| O_t) \tag{mean-field approximation}
\end{align*}
where $\nb{\rho_J^s}$ is the number of parents of $S^s_J$, and $\rho_{J,i}^s$ is the $i$-th parent of $S^s_J$. Importantly, $P(S_J^s| \rho_J^s)$ is known from the definition of the generative model. Moreover, since $\rho_{J,i}^s \in S_I$, then $P(\rho_{J,i}^s| O_t) = P(S_{I}^m|O_t)$ for some $m \in \{1, \hdots, \nb{S}\}$. Thus, $P(\rho_{J,i}^s| O_t)$ is given as input to the P-step, i.e., $P(\rho_{J,i}^s| O_t)$ is a known distribution. Similarly, the computation of  $P(O_{I{::}\bm{a}}^o|O_t)$ proceeds as follows:
\begin{align*}
P(O_{I{::}\bm{a}}^o|O_t) \delequal P(O_J^o|O_t) &= \sum_{\rho_J^o}^{{\color{white}M}} P(O_J^o, \rho_J^o |O_t) \tag{sum rule}\\
&= \sum_{\rho_J^o}^{{\color{white}M}} P(O_J^o| \rho_J^o, O_t)P(\rho_J^o| O_t) \tag{product rule}\\
&= \sum_{\rho_J^o}^{{\color{white}M}} P(O_J^o| \rho_J^o)P(\rho_J^o| O_t) \tag{d-separation}\\
&\approx \sum_{\rho_J^o} P(O_J^o| \rho_J^o) \prod_{i=1}^{\nb{\rho_J^o}} P(\rho_{J,i}^o| O_t) \tag{mean-field approximation}
\end{align*}
where $\nb{\rho_J^o}$ is the number of parents of $O^o_J$, and $\rho_{J,i}^o$ is the $i$-th parent of $O^o_J$. Importantly, $P(O_J^o| \rho_J^o)$ is known from the definition of the generative model. Moreover, since $\rho_{J,i}^o \in S_J$, then $P(\rho_{J,i}^o| O_t) = P(S_{J}^s|O_t)$ for some $s \in \{1, \hdots, \nb{S}\}$. Thus, $P(\rho_{J,i}^o| O_t)$ has already been computed during the first stage of the P-step and is a known distribution, c.f., derivation of $P(S_{I{::}\bm{a}}^s|O_t) \delequal P(S_J^s|O_t)$.

\subsection{Expected Free Energy} \label{ssec:efe}

In this section, we discuss the definition of the expected free energy, which quantifies the cost of pursuing a particular sequence of actions and will be useful for planning, cf. Section \ref{ssec:planning}. The expected free energy (see below) is composed of the risk and ambiguity terms. The risk terms quantify how much the posterior beliefs over future observations (computed by the P-step) diverge from the prior preferences of the agent. On the other hand, the ambiguity terms correspond to the expected uncertainty of the likelihood mapping, where the expectation is with respect to the posterior beliefs over states computed by the P-step.

First, we partition the set of observations $O_I = \{O_I^o \mid o = 1, \hdots, \nb{O}\}$ into disjoint subsets $X_i^I$, i.e., $O_I = X_1^I \cup \hdots \cup X_N^I$ and $X_i^I \cap X_j^I = \emptyset$ if $i \neq j$. Then, we define the prior preferences over the $i$-th subset of observations as: $V(X_i^I) = \text{Cat}(\bm{C}^i)$. This formulation allows us to define prior preferences over subsets of random variables, and will be useful in Section \ref{ssec:dsprites}, where the agent needs to possess preferences that depend upon both the shape and $(X, Y)$ position of the object. Finally, the expected free energy, which needs to be minimised, is given by:
\begin{align}\label{eq:efe}
\bm{G}_I \delequal \sum_{i=1}^{N} \Bigg( \underbrace{D_{\mathrm{KL}}[P(X_i^I|O_t)||V(X_i^I)]}_{\text{risk of } i \text{-th set of observations}}\Bigg)\, +\,\, \sum_{o=1}^{\nb{O}} \Bigg( \underbrace{\mathbb{E}_{P(\rho_I^o|O_t)}[\text{H}[P(O_I^o | \rho_I^o)]]}_{\text{ambiguity of } o \text{-th observation}}\Bigg),
\end{align}
where $P(X_i^I|O_t)$ and $P(\rho_I^o|O_t)$ are the posteriors over the $i$-th subset of observations and the parent of $O_I^o$, respectively, and $P(O_I^o | \rho_I^o)$ is known from the generative model. Assuming a mean-field approximation, those posteriors are given by:
\begin{align*}
P(\rho_I^o|O_t) &\approx \prod_{i = 1}^{\nb{\rho_I^o}} P(\rho_{I,i}^o|O_t)\\
P(X_i^I|O_t) \,\,\, &\approx \prod_{O_I^o \in X_i}^{{\color{white}x}} P(O_I^o|O_t)
\end{align*}
where $P(O_I^o|O_t)$ and $P(\rho_{I,i}^o|O_t)$ are the posteriors over $O_I^o$ and the $i$-th parent of $O_I^o$, respectively. Note, both $P(O_I^o|O_t)$ and $P(\rho_{I,i}^o|O_t)$ were computed during the P-step. The definition of the expected free energy given by \eqref{eq:efe} may not be very intuitive. Fortunatly, the special case where each subset contains a single observation, i.e., $X_o^I = O_I^o$, leads to the following equation:
\begin{align*}
\bm{G}_I \delequal \sum_{o=1}^{\nb{O}} \Bigg( \underbrace{D_{\mathrm{KL}}[P(O_I^o|O_t)||V(O_I^o)]}_{\text{risk of } o \text{-th observation}} \,\, +\,\, \underbrace{\mathbb{E}_{P(\rho_I^o|O_t)}[\text{H}[P(O_I^o | \rho_I^o)]]}_{\text{ambiguity of } o \text{-th observation}}\Bigg),
\end{align*}
which is the summation over all observations $O_I^o$ of the expected free energy of $O_I^o$, i.e., the risk of $O_I^o$ plus the ambiguity of $O_I^o$. Finally, our framework allows to specify prior preferences over only a subset of variables in $O_I$. For example, if a task contains four variables, i.e., $O_I^x$, $O_I^y$, $O_I^{shape}$ and $O_I^{scale}$, but it only makes sense to have preferences over three of them, i.e., $O_I^x$, $O_I^y$ and $O_I^{shape}$, then the prior preference over the fourth variable is set to the posterior over this random variable, i.e., $V(O_I^{shape}) \delequal P(O_I^{shape}|O_t)$. In other words, not having prior preferences over a random variable is viewed by our framework as liking whatever we predict will happen. Effectively, this renders the risk term associated with such variable equal to zero, i.e., 
\begin{align*}
D_{\mathrm{KL}}[P(O_I^{shape}|O_t)||V(O_I^{shape})] = D_{\mathrm{KL}}[P(O_I^{shape}|O_t)||P(O_I^{shape}|O_t))] = 0.
\end{align*}

\subsection{Planning: the MCTS algorithm} \label{ssec:planning}

In this section, we describe the planning algorithm used by $BTAI_{3MF}$. At the beginning of a trial when $t = 0$, the agent is provided with the initial observations $O_0$. The I-step is performed and returns the posterior over all latent states, i.e., $P(S_0^s|O_0)$ for all $s \in \{1, \hdots, \nb{S}\}$, according to the prior over the initial hidden states provided by the modeller, i.e., $P(S_0^s)$ for all $s \in \{1, \hdots, \nb{S}\}$, and the available observations $O_0$. 

Then, we use the UCT criterion to determine which node in the tree should be expanded. Let the tree's root $TS_t$ be called the current node. If the current node has no children, then it is selected for expansion. Alternatively, the child with the highest UCT criterion becomes the new current node and the process is iterated until we reach a leaf node (i.e. a node from which no action has previously been selected). The UCT criterion \citep{MCTS} for the $j$-th child of the current node is given by:
\begin{align}\label{eq:UCT}
UCT_j = - \bar{\bm{G}}_j + C_{explore} \sqrt{\frac{\ln n}{n_j}},
\end{align}
where $\bar{\bm{G}}_j$ is the average expected free energy calculated with respected to the actions selected from the $j$-th child, $C_{explore}$ is the exploration constant that modulates the amount of exploration at the tree level, $n$ is the number of times the current node has been visited, and $n_j$ is the number of times the $j$-th child has been visited. 

Let $S_I$ be the (leaf) node selected by the above selection procedure. We then expand all the children of $S_I$, i.e., all the states of the form $S_{I{::}\bm{a}}$, where $\bm{a} \in \{1, ..., \nb{A}\}$ is an arbitrary action, $\nb{A}$ is the number of available actions, and $I{::}\bm{a}$ is the multi-index obtained by appending the action $\bm{a}$ at the end of the sequence defined by $I$. Next, we perform the P-step for each action $\bm{a}$, and obtain $P(S_{I{::}\bm{a}}^s|O_t)$ for each latent state $s \in \{1, \hdots, \nb{S}\}$ and $P(O_{I{::}\bm{a}}^o|O_t)$ for each future observation $o \in \{1, \hdots, \nb{O}\}$.

Then, we need to estimate the cost of (virtually) taking each possible action. The cost in this paper is taken to be the expected free energy given by \eqref{eq:efe}. Next, we assume that the agent will always perform the action with the lowest cost, and back-propagate the cost of the best (virtual) action toward the root of the tree. Formally, we write the update as follows:
\begin{align}\label{eq:backprop}
\forall K \in \mathbb{A}_I \cup \{I\}, \quad \bm{G}_K \leftarrow \bm{G}_K + \min_{\bm{a} \in \{1, ..., \nb{A}\}} \bm{G}_{I{::}\bm{a}},
\end{align}
where $I$ is the multi-index of the node that was selected for (virtual) expansion, and $\mathbb{A}_I$ is the set of all multi-indices corresponding to ancestors of $TS_I$. During the back propagation, we also update the number of visits as follows:
\begin{align}\label{eq:backprop_n}
\forall K \in \mathbb{A}_I \cup \{I\}, \quad n_K \leftarrow n_K + 1.
\end{align}
If we let $\bm{G}^{aggr}_K$ be the aggregated cost of an arbitrary node $S_K$ obtained by applying Equation \ref{eq:backprop} after each expansion, then we are now able to express $\bar{\bm{G}}_K$ formally as:
$$\bar{\bm{G}}_K = \frac{\bm{G}^{aggr}_K}{n_K}.$$
The planning procedure described above ends when the maximum number of planning iterations is reached.

\subsection{Action selection} \label{ssec:action_selection}

After performing planning, the agent needs to choose the action to perform in the environment. As discussed in Section 3.1 of \citep{MCTS}, many possible mechanisms can be used to select the action to perform in the environment. $BTAI_{3MF}$ performs the action corresponding to the root child with the highest number of visits. Formally, this is expressed as:
\begin{align}\label{eq:action_selection}
\bm{a}^* = \argmax_{\bm{a} \in \{1, ..., \nb{A}\}} n_{(\bm{a})},
\end{align}
where $\bm{a}^*$ is the action performed in the environment, and $n_{(\bm{a})}$ is the number of visits of the root child corresponding to action $\bm{a}$.

\subsection{Closing the action-perception cycle}

After performing an action $\bm{a}^*$ in the environment, the agent receives a new observation $O_{t+1}$, and needs to use this observation to compute the posterior over the latent states at time $t+1$, i.e., $P(S^s_{t+1}|O_{t+1})$ for all $s \in \{1, \hdots, \nb{S}\}$. This can be achieved by performing the I-step, but requires the agent to have prior beliefs over the latent states at time $t+1$, i.e., $P(S^s_{t+1})$ for all $s \in \{1, \hdots, \nb{S}\}$, in addition to the new observation $O_{t+1}$ obtained from the environment. In this paper, we define those prior beliefs as:
\begin{align*}
P(S^s_{t+1}) = P(S_{I}^s|O_t), \text{ for all } s \in \{1, \hdots, \nb{S}\},
\end{align*}
where $I = (\bm{a}^*)$ is a sequence of actions containing the action $\bm{a}^*$ performed in the environment, $P(S_I^s|O_t)$ is the predictive posterior computed by the P-step when assuming that action $\bm{a}^*$ is performed. In other words, the predictive posterior $P(S_I^s|O_t)$ computed by the P-step at time $t$, is used as an empirical prior $P(S^s_{t+1})$ at time $t+1$. This empirical prior $P(S^s_{t+1})$ along with the new observation $O_{t+1}$ can then be used to compute the posterior $P(S^s_{t+1}|O_{t+1})$ for all $s \in \{1, \hdots, \nb{S}\}$. This posterior will be used to perform planning in the next action-perception cycle. Algorithm \ref{algo:BTAI_3MF_cycles} concludes this section by summarizing our approach.

\begin{algorithm}[H]
\label{algo:BTAI_3MF_cycles}
\SetAlgoLined\DontPrintSemicolon
\SetKwInOut{Input}{Input}
\SetKwFor{RepTimes}{repeat}{times}{end}
\SetAlgoLined
\Input{$env$ the environment,\\
$O_0 = \{O^o_0 \mid o = 1, \hdots \nb{O}\}$ the initial observations,\\
$\bm{A} = \{\bm{A}^o \mid o = 1, \hdots \nb{O}\}$ the likelihood mapping of each observation,\\
$\bm{B} = \{\bm{B}^s \mid s = 1, \hdots, \nb{S}\}$ the transition mapping for each hidden state,\\
$\bm{C} = \{\bm{C}^i \mid i = 1, \hdots N\}$ the prior preferences of each subset of observations,\\
$\bm{D}_0 = \{\bm{D}_0^s \mid s = 1, \hdots \nb{S}\}$ the prior over each initial state,\\
$N$ the number of planning iterations,\\
$M$ the number of action-perception cycles.}
 {\color{white}space}\;
 $P(S_0^s|O_0) \leftarrow $ I-step($O_0$, $\bm{A}$, $\bm{D}_0$) \tcp*{I-step from Section \ref{ssec:i_step}}
 $root \leftarrow $ CreateTreeNode(\\
 $\quad$ beliefs = $P(S_0^s|O_0)$, action = -1, cost = 0, visits = 1\\
 )\tcp*{Create the root node for the MCTS, where -1 is a dummy value}
 \RepTimes{$M$} {
 \RepTimes{$N$} {
   $node \leftarrow $ SelectNode($root$) \tcp*{Using (\ref{eq:UCT}) recursively}
   $eNodes \leftarrow $ ExpandChildren($node$, $\bm{B}$) \tcp*{P-step from Section \ref{ssec:p_step} for each action}
   Evaluate($eNodes$, $\bm{A}$, $\bm{C}$) \tcp*{Compute (\ref{eq:efe}) for each expanded node}
   Backpropagate($eNodes$) \tcp*{Using (\ref{eq:backprop}) and (\ref{eq:backprop_n})}
 }
 $\bm{a}^* \leftarrow $ SelectAction($root$) \tcp*{Using \eqref{eq:action_selection}}
 $O_{t+1} \leftarrow $ $env$.Execute($\bm{a}^*$)\;
 $child \leftarrow root.children[\bm{a}^*]$ \tcp*{Get root child corresponding to $\bm{a}^*$}
 $P(S_{t+1}^s) \leftarrow child.beliefs$ \tcp*{Get the empirical prior $P(S^s_{t+1}) = \text{Cat}(\bm{D}_{t+1}^s)$}
 $P(S^s_{t+1}|O_{t+1}) \leftarrow$ I-step($O_{t+1}$, $\bm{A}$, $\bm{D}_{t+1}$) \tcp*{I-step from Section \ref{ssec:i_step}}
 $root \leftarrow $ CreateTreeNode(\\
 $\quad$ beliefs = $P(S^s_{t+1}|O_{t+1})$, action = $\bm{a}^*$, cost = 0, visits = 1\\
 )\tcp*{Create the root node of the next action-perception cycle}
 }
 \caption{$BTAI_{3MF}$: action-perception cycles (with relevant equations indicated in round brackets).}
\end{algorithm}

\section{Results} \label{sec:results}

In this section, we compare our new approach to BTAI with variational message passing ($BTAI_{VMP}$) and BTAI with Bayesian filtering ($BTAI_{BF}$). Section \ref{ssec:dsprites} presents the simplified version of the dSprites environment on which the agents are compared. Section \ref{ssec:btai_vmp} describes how the task is modelled by the $BTAI_{VMP}$ agent and reports its performance, finally, Sections \ref{ssec:btai_bf} and \ref{ssec:btai_3mf} do the same for the $BTAI_{BF}$ and $BTAI_{3MF}$ agents. For the reader interested in implementing a custom $BTAI_{3MF}$ agent, Appendix A provides a tutorial of how to create such an agent using our framework, and Appendix B desbribes a graphical user interface (GUI) that can be used to inspect the model. This GUI displays the structure of the generative model and prior preferences, the posterior beliefs of each latent variable, the messages sent throughout the factor graph to perform inference, the information related to the MCTS algorithm, and the expected free energy (EFE) of each node in the future. It also shows how the EFE decomposes into the risk and ambiguity terms.

\subsection{dSprites Environment} \label{ssec:dsprites}

\begin{figure}[H]
	\begin{center}
	\includegraphics[scale=2]{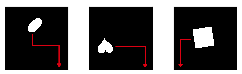}
	\end{center}
  \caption{This figure illustrates the dSprites environment, in which the agent must move all squares towards the bottom-left corner of the image and all ellipses and hearts towards the bottom-right corner of the image. The red arrows show the behaviour expected from the agent.}
   \label{fig:dSprites_env}
\end{figure}

The dSprites environment is based on the dSprites dataset \citep{dsprites17} initially designed for analysing the latent representation learned by variational auto-encoders \citep{VAE}. The dSprites dataset is composed of images of squares, ellipses and hearts. Each image contains one shape (square, ellipse or heart) with its own scale, orientation, and $(X,Y)$ position. In the dSprites environment, the agent is able to move those shapes around by performing four actions (i.e., UP, DOWN, LEFT, RIGHT). To make planning tractable, the action selected by the agent is executed eight times in the environment before the beginning of the next action-perception cycle, i.e., the $X$ or $Y$ position is increased or decreased by eight between time step $t$ and $t+1$. The goal of the agent is to move all squares towards the bottom-left corner of the image and all ellipses and hearts towards the bottom-right corner of the image, c.f. Figure \ref{fig:dSprites_env}.

Since BTAI is a tabular model whose likelihood and transition mappings are represented using matrices, the agent does not directly take images as inputs. Instead, the metadata of the dSprites dataset is used to specify the state space. In particular, the agent observes the type of shape (i.e., square, ellipse, or heart), the scale and orientation of the shape, as well as a coarse-grained version of the shape's true position. Importantly, the original images are composed of 32 possible values for both the $X$ and $Y$ positions of the shapes. A coarse-grained  representation with a granularity of two means that the agent is only able to perceive $16 \times 16$ images, and thus, the positions at coordinate $(0,0)$, $(0,1)$, $(1,0)$ and $(1,1)$ are indistinguishable. Figure \ref{fig:down_sampling} illustrates the coarse grained representation with a granularity of eight and the corresponding indices observed by the $BTAI_{VMP}$ and $BTAI_{BF}$ agents. Note that this modification of the observation space can be seen as a form of state aggregation \citep{STATES_AGGREG}. Finally, as shown in Figure \ref{fig:down_sampling}, the prior preferences of the agent are specified over an absorbing row below the dSprites image. This absorbing row ensures that the agent selects the action ``down" when standing in the ``appropriate corner", i.e., bottom-left corner for squares and bottom-right coner for ellipses and hearts.

{
\colorlet{Green}{green!70!black}
\colorlet{Red}{red!80!black}
\renewcommand\fbox{\fcolorbox{white}{white}}
\begin{figure}[H]
	\begin{center}
	\begin{tikzpicture}[scale=0.4, every node/.style={scale=0.4}]
	\node at (-5.5, 2.5) {\includegraphics[scale=2.5]{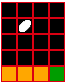}};

	\draw[step=1.0,black,thin] (0,0) grid (4,5);	
	\draw[step=1.0,black,thin] (7,0) grid (11,5);	
	\draw[step=1.0,black,thin] (14,0) grid (18,5);	
	
	\node[scale=2.5] at (2,-1) {$\square$};
	\node[scale=2.5] at (9,-1) {$\heart$};
	\draw (16,-1) ellipse (0.5cm and 0.3cm);

	\node[scale=2] at (0.5,4.5) {0};
	\node[scale=2] at (0.5,3.5) {1};
	\node[scale=2] at (0.5,2.5) {2};
	\node[scale=2] at (0.5,1.5) {3};
	\node[scale=2] at (0.5,0.5) {4};

	\node[scale=2] at (1.5,4.5) {5};
	\node[scale=2] at (1.5,3.5) {6};
	\node[scale=2] at (1.5,2.5) {7};
	\node[scale=2] at (1.5,1.5) {8};
	\node[scale=2] at (1.5,0.5) {9};

	\node[scale=2] at (2.5,4.5) {10};
	\node[scale=2] at (2.5,3.5) {11};
	\node[scale=2] at (2.5,2.5) {12};
	\node[scale=2] at (2.5,1.5) {13};
	\node[scale=2] at (2.5,0.5) {14};

	\node[scale=2] at (3.5,4.5) {15};
	\node[scale=2] at (3.5,3.5) {16};
	\node[scale=2] at (3.5,2.5) {17};
	\node[scale=2] at (3.5,1.5) {18};
	\node[scale=2] at (3.5,0.5) {19};

	\node[scale=2] at (7.5,4.5) {20};
	\node[scale=2] at (7.5,3.5) {21};
	\node[scale=2] at (7.5,2.5) {22};
	\node[scale=2] at (7.5,1.5) {23};
	\node[scale=2] at (7.5,0.5) {24};

	\node[scale=2] at (8.5,4.5) {25};
	\node[scale=2] at (8.5,3.5) {26};
	\node[scale=2] at (8.5,2.5) {27};
	\node[scale=2] at (8.5,1.5) {28};
	\node[scale=2] at (8.5,0.5) {29};

	\node[scale=2] at (9.5,4.5) {30};
	\node[scale=2] at (9.5,3.5) {31};
	\node[scale=2] at (9.5,2.5) {32};
	\node[scale=2] at (9.5,1.5) {33};
	\node[scale=2] at (9.5,0.5) {34};

	\node[scale=2] at (10.5,4.5) {35};
	\node[scale=2] at (10.5,3.5) {36};
	\node[scale=2] at (10.5,2.5) {37};
	\node[scale=2] at (10.5,1.5) {38};
	\node[scale=2] at (10.5,0.5) {39};

	\node[scale=2] at (14.5,4.5) {40};
	\node[scale=2] at (14.5,3.5) {41};
	\node[scale=2] at (14.5,2.5) {42};
	\node[scale=2] at (14.5,1.5) {43};
	\node[scale=2] at (14.5,0.5) {44};
	
	\node[scale=2] at (15.5,4.5) {45};
	\node[scale=2] at (15.5,3.5) {46};
	\node[scale=2] at (15.5,2.5) {47};
	\node[scale=2] at (15.5,1.5) {48};
	\node[scale=2] at (15.5,0.5) {49};
	
	\node[scale=2] at (16.5,4.5) {50};
	\node[scale=2] at (16.5,3.5) {51};
	\node[scale=2] at (16.5,2.5) {52};
	\node[scale=2] at (16.5,1.5) {53};
	\node[scale=2] at (16.5,0.5) {54};
	
	\node[scale=2] at (17.5,4.5) {55};
	\node[scale=2] at (17.5,3.5) {56};
	\node[scale=2] at (17.5,2.5) {57};
	\node[scale=2] at (17.5,1.5) {58};
	\node[scale=2] at (17.5,0.5) {59};
	    \end{tikzpicture}
	\end{center}
\vspace{-0.25cm}
    \caption{
This figure illustrates the observations made by the agent when using a coarse-grained representation with a granularity of eight on the input image. On the left, one can see an image from the dSprites dataset and a grid containing red squares of $8\times8$ pixels. Any positions in those $8\times8$ squares are indistinguishable from the perspective of the agent. Also, the bottom most row is an absorbing row used to specify the prior preferences of the agent, i.e. the green square is the goal state and the orange squares correspond to undesirable states. Finally, the three tables on the right contain the indices observed by the $BTAI_{VMP}$ and $BTAI_{BF}$ agents for each type of shape at each possible position.}
    \label{fig:down_sampling}
\end{figure}
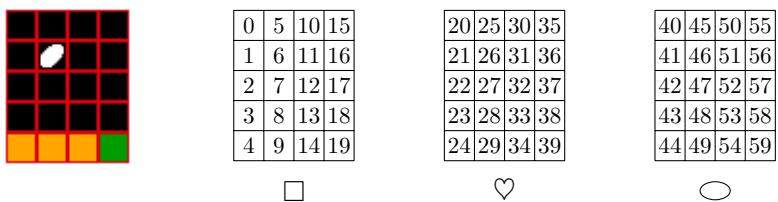
}

The evaluation of the agent's performance is based on the reward obtained by the agent. Briefly, the agent receives a reward of $-1$, if it never enters the absorbing row or if it does so at the antipode of the appropriate corner. As the agent enters the absorbing row closer and closer to the appropriate corner, its reward increases until reaching a maximum of $1$. The percentage of the task solved (i.e., the evaluation metric) is calculated as follows:
$$P(\text{solved}) = \frac{\text{total rewards} + \text{number of runs}}{2.0 \times \text{number of runs}}.$$
Intuitively, the numerator shifts the rewards so that they are bounded between zero and two, and the denominator renormalises the reward to give a score between zero and one. A score of zero therefore corresponds to an agent always failing to enter the absorbing row or doing so at the antipode of the appropriate corner. In contrast, a score of one corresponds to an agent always entering the absorbing row through the appropriate corner.

\subsection{$BTAI_{VMP}$ modeling approach and results} \label{ssec:btai_vmp}

In this section, we evaluate $BTAI_{VMP}$ \citep{AITS_THEORY,AITS_PRACTICE} on the dSprites environment. As shown in Figure \ref{fig:down_sampling}, $BTAI_{VMP}$ observes one index for each possible configuration of shape, and $(X, Y)$ positions. Importantly, this version of BTAI suffers from the exponential growth described in the introduction, and thus does not model the scale and orientation modalities. Also, to make the inference and planning process tractable, the granularity of the coarse-grained representation was set to four or eight. Table \ref{tab:values_hp_BTAI_dSprites} provides the value of each hyper-parameter used by $BTAI_{VMP}$ in this section. Note, the hyper-parameter values are the same for all BTAI models presented in this paper. Only the number of action perception cycles, and the number of planning iterations may vary from one experiment to the next.

\begin{table}[H]
\centering
\begin{tabular}{ |c|c|  }
 \hline
 Name & Value\\
 \hline
 \hline
 \url{NB_SIMULATIONS} & 100\\
 \hline
 \url{NB_ACTION_PERCEPTION_CYCLES} & 30\\
 \hline
 \url{NB_PLANNING_STEPS} & 10, 25 or 50\\
 \hline
 \url{EXPLORATION_CONSTANT} & 2.4\\
 \hline
 \url{PRECISION_PRIOR_PREFERENCES} & 2\\
 \hline
 \url{PRECISION_ACTION_SELECTION} & 100\\
 \hline
 \url{EVALUATION_TYPE} & EFE\\
 \hline
\end{tabular}
\caption{The value of each hyper-parameter used by $BTAI_{VMP}$ in this section. \url{NB_SIMULATIONS} is the number of simulations run during the experiment. \url{NB_ACTION_PERCEPTION_CYCLES} is the maximum number of actions executed in each simulation, after which the simulation is terminated. \url{NB_PLANNING_STEPS} is the number of planning iterations performed by the agent. \url{EXPLORATION_CONSTANT} is the exploration constant of the UCT criterion. \url{PRECISION_PRIOR_PREFERENCES} is the precision of the prior preferences. \url{PRECISION_ACTION_SELECTION} is the precision of the distribution used for action selection. \url{EVALUATION_TYPE} is the type of cost used to evaluate the node during the tree search. Those hyper-parameters can be used to re-run the experiments using the code of the following GitHub repository: \url{https://github.com/ChampiB/Experiments_AI_TS}.}
\label{tab:values_hp_BTAI_dSprites}
\end{table}

Briefly, the agent is able to solve 88.5\% of the task when using a granularity of eight, c.f. Table \ref{tab:dSprites_res_8}. To understand why $BTAI_{VMP}$ was not able to solve the task with 100\% accuracy, let us consider the example of an ellipse at position $(24,31)$. With a granularity of eight, the agent perceives that the ellipse is in the bottom-right corner of the image, i.e., in the red square just above the goal state in Figure \ref{fig:down_sampling}. From the agent's perspective, it is thus optimal to pick the action ``down" to reach the goal state. However, in reality, the agent will not receive the maximum reward because its true $X$ position is $24$ instead of the optimal $X$ position of $31$. 

\begin{table}[H]
\centering
\begin{tabular}{ |c|c|c| }
 \hline
 Planning iterations & P(solved) & Time (sec) \\
 \hline
 10 & 0.813	 & 0.859 $\pm$ 0.868 \\
 \hline
 25 & 0.846 & 0.862 $\pm$ 0.958 \\
 \hline
 50 & 0.885 & 1.286 $\pm$ 1.261 \\
 \hline
\end{tabular}
\caption{The percentage of the dSprites environment solved by the $BTAI_{VMP}$ agent when using a granularity of eight, c.f. Figure \ref{fig:down_sampling}. The last column reports the average execution time required for one simulation and the associated standard deviation.}
\label{tab:dSprites_res_8}
\end{table}

As shown in Table \ref{tab:dSprites_res_4}, we can improve the agent's perfomance, by using a granularity of four. This allows the agent to differentiate between a larger number of $(X,Y)$ positions, i.e., it reduces the size of the red square in Figure \ref{fig:down_sampling}. With this setting, the agent is able to solve 96.9\% of the task. However, when decreasing the granularity, the number of states goes up, and so does the width and height of the $\bm{A}$ and $\bm{B}$ matrices. As a result, more memory and computational time is required for the inference and planning process. This highlights a trade-off between the agent's performance and the amount of memory and time required. Indeed, a smaller granularity leads to better performance, but requires more time and memory.

\begin{table}[H]
\centering
\begin{tabular}{ |c|c|c| }
 \hline
 Planning iterations & P(solved) & Time (sec) \\
 \hline
 10 & 0.859 & 3.957 $\pm$ 4.027 \\
 \hline
 25 & 0.933 & 3.711 $\pm$ 4.625 \\
 \hline
 50 & 0.969 & 5.107 $\pm$ 5.337 \\
 \hline
\end{tabular}
\caption{The percentage of the dSprites environment solved by the $BTAI_{VMP}$ agent when using a granularity of four. In this setting, there are $9 \times 8 \times 3 = 216$ states. The last column reports the average execution time required for one simulation and the associated standard deviation.}
\label{tab:dSprites_res_4}
\end{table}

\subsection{$BTAI_{BF}$ modeling approach and results} \label{ssec:btai_bf}

In this section, we evaluate $BTAI_{BF}$ \citep{BTAI_BF} on the dSprites environment. As shown in Figure \ref{fig:down_sampling}, $BTAI_{BF}$ observes one index for each possible configuration of shape, and $(X, Y)$ positions. Also, to make the inference and planning process tractable, the granularity of the coarse-grained representation was set to two, four or eight. Table \ref{tab:values_hp_BTAI_BF_dSprites} provides the value of each hyper-parameter used by $BTAI_{BF}$ in this section. Note, the hyper-parameter values are the same for all BTAI models presented in this paper. Only the number of action perception cycles, and the number of planning iterations may vary from one experiment to the next.

\begin{table}[H]
\centering
\begin{tabular}{ |c|c|  }
 \hline
 Name & Value\\
 \hline
 \hline
 \url{NB_SIMULATIONS} & 100\\
 \hline
 \url{NB_ACTION_PERCEPTION_CYCLES} & 20\\
 \hline
 \url{NB_PLANNING_STEPS} & 50\\
 \hline
 \url{EXPLORATION_CONSTANT} & 2.4\\
 \hline
 \url{PRECISION_PRIOR_PREFERENCES} & 1\\
 \hline
 \url{PRECISION_ACTION_SELECTION} & 100\\
 \hline
 \url{EVALUATION_TYPE} & EFE\\
 \hline
\end{tabular}
\caption{The value of each hyper-parameter used by $BTAI_{BF}$ in this section. \url{NB_SIMULATIONS} is the number of simulations run during the experiment. \url{NB_ACTION_PERCEPTION_CYCLES} is the maximum number of actions executed in each simulation, after which the simulation is terminated. \url{NB_PLANNING_STEPS} is the number of planning iterations performed by the agent. \url{EXPLORATION_CONSTANT} is the exploration constant of the UCT criterion. \url{PRECISION_PRIOR_PREFERENCES} is the precision of the prior preferences. \url{PRECISION_ACTION_SELECTION} is the precision of the distribution used for action selection. \url{EVALUATION_TYPE} is the type of cost used to evaluate the node during the tree search. Those hyper-parameters can be used to re-run the experiments using the code of the following GitHub repository: \url{https://github.com/ChampiB/Branching_Time_Active_Inference}.}
\label{tab:values_hp_BTAI_BF_dSprites}
\end{table}

As shown in Table \ref{tab:btai_bf_dSprites_res}, the agent is able to solve: 86.1\% of the task when using a granularity of eight, 97.7\% of the task when using a granularity of four, and 98.6\% of the task when using a granularity of two. However, as the performance improves from 86.1\% to 98.6\%, the computational time required to run each simulation skyrockets from around 50 milliseconds to around 17.5 seconds. In other words, a simulation with a granularity of two is 350 times slower than a simulation with a granularity of eight.

\begin{table}[H]
\centering
\begin{tabular}{ |c|c|c|c| }
 \hline
 Planning iterations & Granularity & P(solved) & Time (ms) \\
 \hline
 50 & 8 & 0.861 & 49.93 $\pm$ 36.4124 \\
 \hline
 50 & 4 & 0.977 & 241.63 $\pm$ 118.379 \\
 \hline
 50 & 2 & 0.986 & 17503.8 $\pm$ 12882.8 \\
 \hline
\end{tabular}
\caption{The percentage of the dSprites environment solved by the $BTAI_{BF}$ agent when using a granularity of eight, four and two. Note, when a granularity of two is used, there are $17 \times 16 \times 3 = 816$ possible states. The last column reports the average execution time required for one simulation and the associated standard deviation. Note, the change in time granularity to milliseconds.}
\label{tab:btai_bf_dSprites_res}
\end{table}

\subsection{$BTAI_{3MF}$ modeling approach and results} \label{ssec:btai_3mf}

In this section, we evaluate our new approach ($BTAI_{3MF}$) on the dSprites environment. In contrast to what is shown in Figure \ref{fig:down_sampling}, $BTAI_{3MF}$ does not observe one index for each possible configuration of shape, and $(X, Y)$ positions. Instead, $BTAI_{3MF}$ has five observed variables representing the shape, the orientation, the scale, as well as the X and Y position, respectively. Each of those observed variable has its hidden state counterparts. Each observation depends on its hidden state counterparts through an identity matrix. This parametrisation is common in the literature on active inference, see \citep{10.1162/neco_a_01357} for an example. The transition mappings of the hidden variables representing the shape, orientation, and scale, are defined as an indentity matrix. This forwards the state value at time $t$ to the next time step $t + 1$. For the hidden variables representing the X and Y position of the shape, the transition is set to reflect the dynamics of the dSprites environment when the actions taken are repeated eight times, i.e., if the action ``DOWN" is selected, then the agent's position in Y will be decreased by eight before the start of the next action-perception cycle \citep{DeepAIwithMCMC}.

The hyper-parameters used in those simualtions are presented in Table \ref{tab:values_hp_BTAI_3MF_dSprites}. Note, the hyper-parameter values are the same for all BTAI models presented in this paper. Only the number of action perception cycles, and the number of planning iterations may vary from one experiment to the next.

Table \ref{tab:btai_3mf_dSprites_res} shows the results obtained by $BTAI_{3MF}$ on the dSprites environment when running 100 trials. Due to the change in the format of representations, the agent exhibits little increase in execution time as the granularity decreases, however, in general, the capacity to solve the task increases with this reduction in granularity. When a granularity of one is used, the agent is able to solve the task perfectly with 150 planning iterations.

Note, the agent using a granularity of 1 and 150 planning iterations is as fast as the agent using a granularity of 1 and 50 planning iterations. This is because as the number of planning iterations increases the agent requires more computation time per action-perception cycle, but as the agent performance increases on the task, the agent reaches the goal state faster, and therefore requires less action-perception cycles per simulation. To conclude, the agent with 150 planning iterations requires less action-perception cycles per simulation, but more time per action-perception cycle than the agent with 50 planning iterations. The code relevant to this section is available at the following URL: \url{https://github.com/ChampiB/BTAI_3MF}.

\begin{table}[H]
\centering
\begin{tabular}{ |c|c|  }
 \hline
 Name & Value\\
 \hline
 \hline
 \url{NB_SIMULATIONS} & 100\\
 \hline
 \url{NB_ACTION_PERCEPTION_CYCLES} & 50\\
 \hline
 \url{NB_PLANNING_STEPS} & 50 or 100 or 150\\
 \hline
 \url{EXPLORATION_CONSTANT} & 2.4\\
 \hline
 \url{PRECISION_PRIOR_PREFERENCES} & 1\\
 \hline
 \url{EVALUATION_TYPE} & EFE\\
 \hline
\end{tabular}
\caption{The value of each hyper-parameter used by $BTAI_{3MF}$ in this section. \url{NB_SIMULATIONS} is the number of simulations run during the experiment. \url{NB_ACTION_PERCEPTION_CYCLES} is the maximum number of actions executed in each simulation, after which the simulation is terminated. \url{NB_PLANNING_STEPS} is the number of planning iterations performed by the agent. \url{EXPLORATION_CONSTANT} is the exploration constant of the UCT criterion. \url{PRECISION_PRIOR_PREFERENCES} is the precision of the prior preferences. \url{EVALUATION_TYPE} is the type of cost used to evaluate the node during the tree search. Those hyper-parameters can be used to re-run the experiments using the code of the following GitHub repository: \url{https://github.com/ChampiB/BTAI_3MF}.}
\label{tab:values_hp_BTAI_3MF_dSprites}
\end{table}

\begin{table}[H]
\centering
\begin{tabular}{ |c|c|c|c| }
 \hline
 Planning iterations & Granularity & P(solved) & Time (sec) \\
 \hline
 50 & 8 & 0.895 & 1.279 $\pm$ 12.8 \\
 \hline
 50 & 4 & 0.977 & 1.279 $\pm$ 12.8 \\
 \hline
 50 & 2 & 0.996 & 1.279 $\pm$ 12.8 \\
 \hline
 50 & 1 & 0.72 & 2.559 $\pm$ 18.01 \\
 \hline
 100 & 1 & 0.77 & 5.119 $\pm$ 25.209 \\
 \hline
 150 & 1 & 1 & 2.559 $\pm$ 18.01 \\
 \hline
\end{tabular}
\caption{This table presents the percentage of the dSprites environment solved by the $BTAI_{3MF}$ agent when using a granularity of eight, four, two and one. Note, when a granularity of one is used, there are $33 \times 32 \times 3 \times 40 \times 6 = 760,320$ possible state configurations. The last column reports the average execution time required of one simulation and the associated standard deviation.}
\label{tab:btai_3mf_dSprites_res}
\end{table}

\section{Conclusion} \label{sec:conclusion}

In this paper, we presented a new version of Branching Time Active Inference that allows for modelling of several observed and latent variables. Taken together, those variables constitute a temporal slice. Within a slice, the model is equipped with prior beliefs over the initial latent variables, and each observation depends on a subset of the latent variables through the likelihood mapping. Additionally, the latent states evolve over time according to the transition mapping that describes how each latent variable at time $t+1$ is generated from a subset of the hidden states at time $t$ and the action taken.

At the beginning of each trial, the agent makes an observation for each observed variable, and computes the posterior over the latent variables using belief propagation. Then, a Monte-Carlo tree search is performed to explore the space of possible policies. During the tree search, each planning iteration starts by selecting a node to expand using the UCT criterion. Then, the children of the selected node are expanded, i.e., one child per action. Next, the posterior over the latent variables of the expanded nodes is computed by performing forward predictions using the known transition mapping, and the posterior beliefs over the latent states of the node selected for expansion. Once the posterior is computed, the expected free energy can be computed and back-propagated through the tree. The planning process stops after reaching a maximum number of iterations.

In the results section, we compared our new approach, called $BTAI_{3MF}$, to two earlier versions of branching time active inference, named $BTAI_{VMP}$ \citep{AITS_THEORY,AITS_PRACTICE} and $BTAI_{BF}$ \citep{BTAI_BF}. Briefly, at the current time step $t$: $BTAI_{VMP}$ performs variational message passing (VMP) with a variational distribution composed of only one factor, $BTAI_{BF}$ performs exact inference using Bayes theorem, and $BTAI_{3MF}$ implements belief propagation to compute the marginal posterior over each latent variable. For the hidden variables in the future, $BTAI_{VMP}$ does the same mean-field approximation as at time step $t$ and performs VMP, $BTAI_{BF}$ performs Bayesian prediction to compute the posterior over the only latent variable being modelled, and likewise, $BTAI_{3MF}$ performs prediction to compute the posterior over all future latent variables.

Since, none of the aforementioned approaches are equipped with deep neural networks, we compared them on a version of the dSprites environment in which the metadata of the dSprites dataset are used as inputs to the model instead of the dSprites images. The best performance obtained by $BTAI_{VMP}$ was to solve 96.9\% of the task in 5.1 seconds. Importantly, $BTAI_{VMP}$  was previously compared to active inference as implemented in SPM both theoretically and experimentally \citep{AITS_THEORY, AITS_PRACTICE}. $BTAI_{BF}$ was able to solve 98.6\% of the task but at the cost of 17.5 seconds of computation. Note, $BTAI_{BF}$ was using a granularity of two (i.e., 816 states) while $BTAI_{VMP}$ was using a granularity of four (i.e., 216 states), which is why $BTAI_{BF}$ seems to be three times slower than $BTAI_{VMP}$. In reality, if $BTAI_{BF}$ had been using a granularity of four, it would have been much faster than $BTAI_{VMP}$ while maintaining a similar performance, i.e., around 96.9\% of the task solved. Finally, $BTAI_{3MF}$ outperformed both of its predecessors by solving the task completely (100\%, granularity of 1) in only 2.559 seconds. Importantly, $BTAI_{3MF}$ was able to model all the modalities of the dSprites environment for a total of $760,320$ possible states.

In addition to the major boost in performance and computational time, $BTAI_{3MF}$ provides an improved modelling capacity. Indeed, the framework can now handle the modelling of several observed and latent variables, and takes advantage of the factorisation of the generative model to perform inference efficiently. As described in detail in Appendix A, we also provide a high level notation for the creation of $BTAI_{3MF}$ that aims to make our approach as staightforward as possible to apply to new domains. The high-level notational language allows the user to create models by simply declaring the variables it contains, and the dependencies between those variables. Then, the framework performs the inference process automatically. Moreover, driven by the need for interpretability, we developed a graphical user interface to analyse the behaviour and reasoning of our agent, which is described in Appendix B.

There are two major directions of future research that may be explored to keep scaling up this framework. First, $BTAI_{3MF}$ is not yet equipped with deep neural networks (DNNs), and is therefore unable to handle certain types of inputs, such as images. In addition to the integration of DNNs into the framework, further research should be performed in order to learn useful sequences of actions. Typically, in the current version of $BTAI_{3MF}$, we built in the fact that each action should be repeated eight times in a row. This inductive bias works well in the context of the dSprites environment, but may be a limitation in other contexts.

It is also worth reflecting on how the $BTAI_{3MF}$ model sits with theories of brain function. In this respect, it is interesting to consider neural correlates of the ``standard" approach that $BTAI_{3MF}$ is being placed in opposition to. As previously discussed, this standard active inference approach could be considered as monolithically tabular; that is, the key matrices, such as the likelihood mapping (the $\bm{A}$ matrix) and the transition mapping (the $\bm{B}$ matrix), grow in size exponentially with the number of states and observations. This is simply due to a combinatorial explosion, e.g. the set of all combinations of states grows intractably with the number of states.

How would the combinations of states in the monolithic tabular approach be represented in the brain? The obvious neural correlate would be conjunctive (binding) neurons \citep{o2001conjunctive}, which become active when multiple feature values are present; for example, one might have a neural unit for every X, Y combination in the dSprites environment. If this is to be realised with a fully localist code, i.e. one unit for every combination, in the absence of any hierarchical structure, the required number of conjunctive units would explode in the same way as the $\bm{A}$ and $\bm{B}$ matrices do. This is why some models have proposed a binding resource that supports distributed (rather than localist) representations \citep{bowman2007simultaneous}, which scale more tractably.

$BTAI_{3MF}$ avoids this combinatorial explosion by not combining features, enabling them to be represented separately. In a very basic sense, this separated representation is consistent with the observation that the brain contains distinct, physically separated, feature maps, e.g. \citet{730558}. Thus, at least to some extent, different feature dimensions are processed separately in the brain, as they are in $BTAI_{3MF}$.

The time-slice idea in $BTAI_{3MF}$ assumes a kind of discrete synchronising global clock. Thus, even though features have been separated from one another and may be considered to execute in different parts of the system, they update in lock-step. That is, implicitly, time is a binder, it determines which values of different feature dimensions/states are associated, e.g. an X-dimension value is associated with a particular Y-dimension value because they are so assigned in the same temporal slice. In this sense, in $BTAI_{3MF}$, time synchronisation resolves the binding problem.

This aspect of $BTAI_{3MF}$ resonates with theories of binding based upon oscillatory synchrony \citep{uhlhaas2009neural}. These theories suggest that different feature dimensions are bound by the corresponding neurons firing in synchrony relative to an ongoing oscillation, with that ongoing oscillation potentially playing the role of a global clock. Such oscillatory synchrony can be seen as a way to resolve the binding problem that does not require conjunctive units.

Conjunction error experiments, e.g. \citet{PMID:11766936}, are also relevant here. In these experiments, participants make errors in associating multiple feature dimensions, perceiving illusory percepts, e.g. if a red K is presented before a blue A in a rapid serial visual presentation stream, in some cases, a red A and a blue K is perceived. These experiments firstly, re-emphasize that different feature dimensions are processed separately, as per $BTAI_{3MF}$: if feature dimensions were not separated, then conjunction errors could not happen. Additionally though, these experiments suggest that there is not a ``perfect" synchronising global clock, since if there were, there would not be any conjunction errors even despite separation of feature dimensions. Generating such conjunction error patterns is an interesting topic for future $BTAI_{3MF}$ modelling work.

\acks{TO BE FILLED}

\vskip 0.2in
\bibliography{references}

\section*{Appendix A: How to create a $BTAI_{3MF}$ agent?}

In this appendix, we describe how to build a $BTAI_{3MF}$ agent using our framework. The relevant code can be found in the file \url{main_BTAI_3MF.py} at the following URL: \url{https://github.com/ChampiB/BTAI_3MF}. Any script running a $BTAI_{3MF}$ agent must start by instantiating an environment in which the agent will be run. Our code provides an implementation of the dSprites environment, which can be created as follow:
\begin{Verbatim}[commandchars=\\\{\}]
\PYG{c+c1}{\PYGZsh{} Create the environment.}
\PYG{n}{env} \PYG{o}{=} \PYG{n}{dSpritesEnv}\PYG{p}{(}\PYG{n}{granularity}\PYG{o}{=}\PYG{l+m+mi}{1}\PYG{p}{,} \PYG{n}{repeat}\PYG{o}{=}\PYG{l+m+mi}{8}\PYG{p}{)}
\PYG{n}{env} \PYG{o}{=} \PYG{n}{dSpritesPreProcessingWrapper}\PYG{p}{(}\PYG{n}{env}\PYG{p}{)}
\end{Verbatim}
The first line creates the dSprites environment, the second makes sure that the observations generated by the environment are in the format expected by the agent. Once the environment has been created, we need to define the parameters of the model. Assume that we want to have a latent variable $S_t^{shape}$ representing the shape in the current image. This variable can takes three values, i.e., zero for squares, one for ellipses and two for hearts. In this case, the parameters of the prior over $S_t^{shape}$ may be created as:
\begin{Verbatim}[commandchars=\\\{\}]
\PYG{c+c1}{\PYGZsh{} Create the parameters of the prior over the latent variable shape.}
\PYG{n}{d} \PYG{o}{=} \PYG{p}{\PYGZob{}\PYGZcb{}}
\PYG{n}{d}\PYG{p}{[}\PYG{l+s+s2}{\PYGZdq{}S\PYGZus{}shape\PYGZdq{}}\PYG{p}{]} \PYG{o}{=} \PYG{n}{torch}\PYG{o}{.}\PYG{n}{tensor}\PYG{p}{([}\PYG{l+m+mf}{0.2}\PYG{p}{,} \PYG{l+m+mf}{0.3}\PYG{p}{,} \PYG{l+m+mf}{0.5}\PYG{p}{])}
\end{Verbatim}
The first line above creates a python dictionary, the second line adds a vector of parameters in the dictionary. This vector can be accessed using the key ``S\_shape", which corresponds to the name of the latent variable. The values in d[``S\_shape"] mean that a priori the agent believes it will observe a square with probability 0.2, an ellipse with probability 0.3, and a heart with probability 0.5. Also, by convention, the name of a latent variable must start with ``S\_". Similarly, if we assume that the shape is provided to the agent through an observed variable $O_t^{shape}$, we can create the parameters of the likelihood mapping for this variable as:
\begin{Verbatim}[commandchars=\\\{\}]
\PYG{c+c1}{\PYGZsh{} Create the parameters of the likelihood mapping for the shape variable.}
\PYG{n}{a} \PYG{o}{=} \PYG{p}{\PYGZob{}\PYGZcb{}}
\PYG{n}{a}\PYG{p}{[}\PYG{l+s+s2}{\PYGZdq{}O\PYGZus{}shape\PYGZdq{}}\PYG{p}{]} \PYG{o}{=} \PYG{n}{torch}\PYG{o}{.}\PYG{n}{eye}\PYG{p}{(}\PYG{l+m+mi}{3}\PYG{p}{)}
\end{Verbatim}
The first line above creates a python dictionary, and the second line adds a 3$\times$3 identity matrix\footnote{Note, in practice the identity matrix is noisy to avoid taking the logarithm of zero.} in the dictionary. This reflects the fact that there is a one-to-one relationship between the value taken by $S_t^{shape}$ and $O_t^{shape}$. Also, by convention, the observations name must start with ``O\_". Since, defining all the parameters manually can be tedious, our framework provides built-in functions that return the model parameters for the dSprites environment. Using those functions, the parameters can be retrieved as follows:
\begin{Verbatim}[commandchars=\\\{\}]
\PYG{c+c1}{\PYGZsh{} Define the parameters of the generative model.}
\PYG{n}{a} \PYG{o}{=} \PYG{n}{env}\PYG{o}{.}\PYG{n}{a}\PYG{p}{()}
\PYG{n}{b} \PYG{o}{=} \PYG{n}{env}\PYG{o}{.}\PYG{n}{b}\PYG{p}{()}
\PYG{n}{c} \PYG{o}{=} \PYG{n}{env}\PYG{o}{.}\PYG{n}{c}\PYG{p}{()}
\PYG{n}{d} \PYG{o}{=} \PYG{n}{env}\PYG{o}{.}\PYG{n}{d}\PYG{p}{(}\PYG{n}{uniform}\PYG{o}{=}\PYG{n+nb+bp}{True}\PYG{p}{)}
\end{Verbatim}
Once all the parameters have been created, it is time to define the structure of the generative model. This can be done using a temporal slice builder, which is an object used to facilitate the creation of a temporal slice. First, we need to create the builder as follows:
\begin{Verbatim}[commandchars=\\\{\}]
\PYG{c+c1}{\PYGZsh{} Create the temporal slice builder.}
\PYG{n}{ts\PYGZus{}builder} \PYG{o}{=} \PYG{n}{TemporalSliceBuilder}\PYG{p}{(}\PYG{l+s+s2}{\PYGZdq{}A\PYGZus{}1\PYGZdq{}}\PYG{p}{,} \PYG{n}{env}\PYG{o}{.}\PYG{n}{n\PYGZus{}actions}\PYG{p}{)}
\end{Verbatim}
The builder takes two parameters, i.e., the name of the action random variable (i.e., ``A\_1") that must start by ``A\_", and the number of possible actions (i.e., env.n\_actions = 4). Then, we need to tell the builder what state variables should be created, and what are the parameters of the prior beliefs over those variables. For the dSprites environment, this can be done as follows:
\begin{Verbatim}[commandchars=\\\{\}]
\PYG{c+c1}{\PYGZsh{} Add the latent states of the model to the temporal slice.}
\PYG{n}{ts\PYGZus{}builder}\PYG{o}{.}\PYG{n}{add\PYGZus{}state}\PYG{p}{(}\PYG{l+s+s2}{\PYGZdq{}S\PYGZus{}pos\PYGZus{}x\PYGZdq{}}\PYG{p}{,} \PYG{n}{d}\PYG{p}{[}\PYG{l+s+s2}{\PYGZdq{}S\PYGZus{}pos\PYGZus{}x\PYGZdq{}}\PYG{p}{])} \PYGZbs{}
    \PYG{o}{.}\PYG{n}{add\PYGZus{}state}\PYG{p}{(}\PYG{l+s+s2}{\PYGZdq{}S\PYGZus{}pos\PYGZus{}y\PYGZdq{}}\PYG{p}{,} \PYG{n}{d}\PYG{p}{[}\PYG{l+s+s2}{\PYGZdq{}S\PYGZus{}pos\PYGZus{}y\PYGZdq{}}\PYG{p}{])} \PYGZbs{}
    \PYG{o}{.}\PYG{n}{add\PYGZus{}state}\PYG{p}{(}\PYG{l+s+s2}{\PYGZdq{}S\PYGZus{}shape\PYGZdq{}}\PYG{p}{,} \PYG{n}{d}\PYG{p}{[}\PYG{l+s+s2}{\PYGZdq{}S\PYGZus{}shape\PYGZdq{}}\PYG{p}{])} \PYGZbs{}
    \PYG{o}{.}\PYG{n}{add\PYGZus{}state}\PYG{p}{(}\PYG{l+s+s2}{\PYGZdq{}S\PYGZus{}scale\PYGZdq{}}\PYG{p}{,} \PYG{n}{d}\PYG{p}{[}\PYG{l+s+s2}{\PYGZdq{}S\PYGZus{}scale\PYGZdq{}}\PYG{p}{])} \PYGZbs{}
    \PYG{o}{.}\PYG{n}{add\PYGZus{}state}\PYG{p}{(}\PYG{l+s+s2}{\PYGZdq{}S\PYGZus{}orientation\PYGZdq{}}\PYG{p}{,} \PYG{n}{d}\PYG{p}{[}\PYG{l+s+s2}{\PYGZdq{}S\PYGZus{}orientation\PYGZdq{}}\PYG{p}{])}
\end{Verbatim}
The function ``add\_state" adds a state variable to the temporal slice. The first parameter of this function is the name of the state to be added, and the second argument is the parameters of the prior beliefs over this new state. Next, we need to add the variables corresponding to the observations made by the agent. For the dSprites environment, this can be done as follows:
\begin{Verbatim}[commandchars=\\\{\}]
\PYG{c+c1}{\PYGZsh{} Define the likelihood mapping of the temporal slice.}
\PYG{n}{ts\PYGZus{}builder}\PYG{o}{.}\PYG{n}{add\PYGZus{}observation}\PYG{p}{(}\PYG{l+s+s2}{\PYGZdq{}O\PYGZus{}pos\PYGZus{}x\PYGZdq{}}\PYG{p}{,} \PYG{n}{a}\PYG{p}{[}\PYG{l+s+s2}{\PYGZdq{}O\PYGZus{}pos\PYGZus{}x\PYGZdq{}}\PYG{p}{],} \PYG{p}{[}\PYG{l+s+s2}{\PYGZdq{}S\PYGZus{}pos\PYGZus{}x\PYGZdq{}}\PYG{p}{])} \PYGZbs{}
    \PYG{o}{.}\PYG{n}{add\PYGZus{}observation}\PYG{p}{(}\PYG{l+s+s2}{\PYGZdq{}O\PYGZus{}pos\PYGZus{}y\PYGZdq{}}\PYG{p}{,} \PYG{n}{a}\PYG{p}{[}\PYG{l+s+s2}{\PYGZdq{}O\PYGZus{}pos\PYGZus{}y\PYGZdq{}}\PYG{p}{],} \PYG{p}{[}\PYG{l+s+s2}{\PYGZdq{}S\PYGZus{}pos\PYGZus{}y\PYGZdq{}}\PYG{p}{])} \PYGZbs{}
    \PYG{o}{.}\PYG{n}{add\PYGZus{}observation}\PYG{p}{(}\PYG{l+s+s2}{\PYGZdq{}O\PYGZus{}shape\PYGZdq{}}\PYG{p}{,} \PYG{n}{a}\PYG{p}{[}\PYG{l+s+s2}{\PYGZdq{}O\PYGZus{}shape\PYGZdq{}}\PYG{p}{],} \PYG{p}{[}\PYG{l+s+s2}{\PYGZdq{}S\PYGZus{}shape\PYGZdq{}}\PYG{p}{])} \PYGZbs{}
    \PYG{o}{.}\PYG{n}{add\PYGZus{}observation}\PYG{p}{(}\PYG{l+s+s2}{\PYGZdq{}O\PYGZus{}scale\PYGZdq{}}\PYG{p}{,} \PYG{n}{a}\PYG{p}{[}\PYG{l+s+s2}{\PYGZdq{}O\PYGZus{}scale\PYGZdq{}}\PYG{p}{],} \PYG{p}{[}\PYG{l+s+s2}{\PYGZdq{}S\PYGZus{}scale\PYGZdq{}}\PYG{p}{])} \PYGZbs{}
    \PYG{o}{.}\PYG{n}{add\PYGZus{}observation}\PYG{p}{(}\PYG{l+s+s2}{\PYGZdq{}O\PYGZus{}orientation\PYGZdq{}}\PYG{p}{,} \PYG{n}{a}\PYG{p}{[}\PYG{l+s+s2}{\PYGZdq{}O\PYGZus{}orientation\PYGZdq{}}\PYG{p}{],} \PYG{p}{[}\PYG{l+s+s2}{\PYGZdq{}S\PYGZus{}orientation\PYGZdq{}}\PYG{p}{])}
\end{Verbatim}
The function ``add\_observation" adds an observation variable to the temporal slice. The first parameter of this function is the name of the observation to be added, the second argument is the parameters of the likelihood mapping for this new observation, and the third parameter is the list of parents on which the observation depends. The next step is the definition of the transition mapping for each hidden state, which can be performed as follows:
\begin{Verbatim}[commandchars=\\\{\}]
\PYG{c+c1}{\PYGZsh{} Define the transition mapping of the temporal slice.}
\PYG{n}{ts\PYGZus{}builder}\PYG{o}{.}\PYG{n}{add\PYGZus{}transition}\PYG{p}{(}\PYG{l+s+s2}{\PYGZdq{}S\PYGZus{}pos\PYGZus{}x\PYGZdq{}}\PYG{p}{,} \PYG{n}{b}\PYG{p}{[}\PYG{l+s+s2}{\PYGZdq{}S\PYGZus{}pos\PYGZus{}x\PYGZdq{}}\PYG{p}{],} \PYG{p}{[}\PYG{l+s+s2}{\PYGZdq{}S\PYGZus{}pos\PYGZus{}x\PYGZdq{}}\PYG{p}{,} \PYG{l+s+s2}{\PYGZdq{}A\PYGZus{}1\PYGZdq{}}\PYG{p}{])} \PYGZbs{}
    \PYG{o}{.}\PYG{n}{add\PYGZus{}transition}\PYG{p}{(}\PYG{l+s+s2}{\PYGZdq{}S\PYGZus{}pos\PYGZus{}y\PYGZdq{}}\PYG{p}{,} \PYG{n}{b}\PYG{p}{[}\PYG{l+s+s2}{\PYGZdq{}S\PYGZus{}pos\PYGZus{}y\PYGZdq{}}\PYG{p}{],} \PYG{p}{[}\PYG{l+s+s2}{\PYGZdq{}S\PYGZus{}pos\PYGZus{}y\PYGZdq{}}\PYG{p}{,} \PYG{l+s+s2}{\PYGZdq{}A\PYGZus{}1\PYGZdq{}}\PYG{p}{])} \PYGZbs{}
    \PYG{o}{.}\PYG{n}{add\PYGZus{}transition}\PYG{p}{(}\PYG{l+s+s2}{\PYGZdq{}S\PYGZus{}shape\PYGZdq{}}\PYG{p}{,} \PYG{n}{b}\PYG{p}{[}\PYG{l+s+s2}{\PYGZdq{}S\PYGZus{}shape\PYGZdq{}}\PYG{p}{],} \PYG{p}{[}\PYG{l+s+s2}{\PYGZdq{}S\PYGZus{}shape\PYGZdq{}}\PYG{p}{])} \PYGZbs{}
    \PYG{o}{.}\PYG{n}{add\PYGZus{}transition}\PYG{p}{(}\PYG{l+s+s2}{\PYGZdq{}S\PYGZus{}scale\PYGZdq{}}\PYG{p}{,} \PYG{n}{b}\PYG{p}{[}\PYG{l+s+s2}{\PYGZdq{}S\PYGZus{}scale\PYGZdq{}}\PYG{p}{],} \PYG{p}{[}\PYG{l+s+s2}{\PYGZdq{}S\PYGZus{}scale\PYGZdq{}}\PYG{p}{])} \PYGZbs{}
    \PYG{o}{.}\PYG{n}{add\PYGZus{}transition}\PYG{p}{(}\PYG{l+s+s2}{\PYGZdq{}S\PYGZus{}orientation\PYGZdq{}}\PYG{p}{,} \PYG{n}{b}\PYG{p}{[}\PYG{l+s+s2}{\PYGZdq{}S\PYGZus{}orientation\PYGZdq{}}\PYG{p}{],} \PYG{p}{[}\PYG{l+s+s2}{\PYGZdq{}S\PYGZus{}orientation\PYGZdq{}}\PYG{p}{])}
\end{Verbatim}
The function ``add\_transition" adds a transition mapping to the temporal slice. The first parameter of this function is the name of the state for which the transition is defined, the second argument is the parameters of the transition mapping for this state, and the third parameter is the list of parents on which the state depends. Importantly, in the above snippet of code, only the states representing the position in x and y of the shape depends on the action variable ``A\_1". The final step is about the defintion of the prior preferences of the agent, and can be done as follows:
\begin{Verbatim}[commandchars=\\\{\}]
\PYG{c+c1}{\PYGZsh{} Define the prior preferences of the temporal slice.}
\PYG{n}{ts\PYGZus{}builder}\PYG{o}{.}\PYG{n}{add\PYGZus{}preference}\PYG{p}{([}\PYG{l+s+s2}{\PYGZdq{}O\PYGZus{}pos\PYGZus{}x\PYGZdq{}}\PYG{p}{,} \PYG{l+s+s2}{\PYGZdq{}O\PYGZus{}pos\PYGZus{}y\PYGZdq{}}\PYG{p}{,} \PYG{l+s+s2}{\PYGZdq{}O\PYGZus{}shape\PYGZdq{}}\PYG{p}{],} \PYG{n}{c}\PYG{p}{[}\PYG{l+s+s2}{\PYGZdq{}O\PYGZus{}shape\PYGZus{}pos\PYGZus{}x\PYGZus{}y\PYGZdq{}}\PYG{p}{])}
\end{Verbatim}
The function ``add\_preference" adds some prior preferences to the temporal slice. The first parameter of this function is the list of observations for which the prior preferences are defined, and the second argument are the parameters of the prior preferences for those observations. At this stage, the initial temporal slice can be built:
\begin{Verbatim}[commandchars=\\\{\}]
\PYG{c+c1}{\PYGZsh{} Create the initial temporal slice.}
\PYG{n}{ts} \PYG{o}{=} \PYG{n}{ts\PYGZus{}builder}\PYG{o}{.}\PYG{n}{build}\PYG{p}{()}
\end{Verbatim}
Once the initial temporal slice has been created, it is possible to instantiate the agent and implement the action-perception cycle as follows: 
\begin{Verbatim}[commandchars=\\\{\}]
\PYG{c+c1}{\PYGZsh{} Create the agent.}
\PYG{n}{agent} \PYG{o}{=} \PYG{n}{BTAI\PYGZus{}3MF}\PYG{p}{(}\PYG{n}{ts}\PYG{p}{,} \PYG{n}{max\PYGZus{}planning\PYGZus{}steps}\PYG{o}{=}\PYG{l+m+mi}{150}\PYG{p}{,} \PYG{n}{exp\PYGZus{}const}\PYG{o}{=}\PYG{l+m+mf}{2.4}\PYG{p}{)}

\PYG{c+c1}{\PYGZsh{} Implement the action\PYGZhy{}perception cycles.}
\PYG{n}{n\PYGZus{}trials} \PYG{o}{=} \PYG{l+m+mi}{100}
\PYG{k}{for} \PYG{n}{i} \PYG{o+ow}{in} \PYG{n+nb}{range}\PYG{p}{(}\PYG{n}{n\PYGZus{}trials}\PYG{p}{):}
    \PYG{n}{obs} \PYG{o}{=} \PYG{n}{env}\PYG{o}{.}\PYG{n}{reset}\PYG{p}{()}
    \PYG{n}{env}\PYG{o}{.}\PYG{n}{render}\PYG{p}{()}
    \PYG{n}{agent}\PYG{o}{.}\PYG{n}{reset}\PYG{p}{(}\PYG{n}{obs}\PYG{p}{)}
    \PYG{k}{while} \PYG{o+ow}{not} \PYG{n}{env}\PYG{o}{.}\PYG{n}{done}\PYG{p}{():}
        \PYG{n}{action} \PYG{o}{=} \PYG{n}{agent}\PYG{o}{.}\PYG{n}{step}\PYG{p}{()}
        \PYG{n}{obs} \PYG{o}{=} \PYG{n}{env}\PYG{o}{.}\PYG{n}{execute}\PYG{p}{(}\PYG{n}{action}\PYG{p}{)}
        \PYG{n}{env}\PYG{o}{.}\PYG{n}{render}\PYG{p}{()}
        \PYG{n}{agent}\PYG{o}{.}\PYG{n}{update}\PYG{p}{(}\PYG{n}{action}\PYG{p}{,} \PYG{n}{obs}\PYG{p}{)}
\end{Verbatim}
Most of the above code is self explanatory. Put simply, this code runs ``n\_trials" simulations of the dSprites environment. The line ``action = agent.step()" performs inference, planning and action selection. The line ``obs = env.execute(action)" executes the selected action in the environment, and the line ``agent.update(action, obs)" updates the agent so that it has taken into account the action taken in the environment and the observations received.

\section*{Appendix B: How to inspect a $BTAI_{3MF}$ agent?}

\begin{figure}[H]
	\begin{center}
	\includegraphics[scale=0.4]{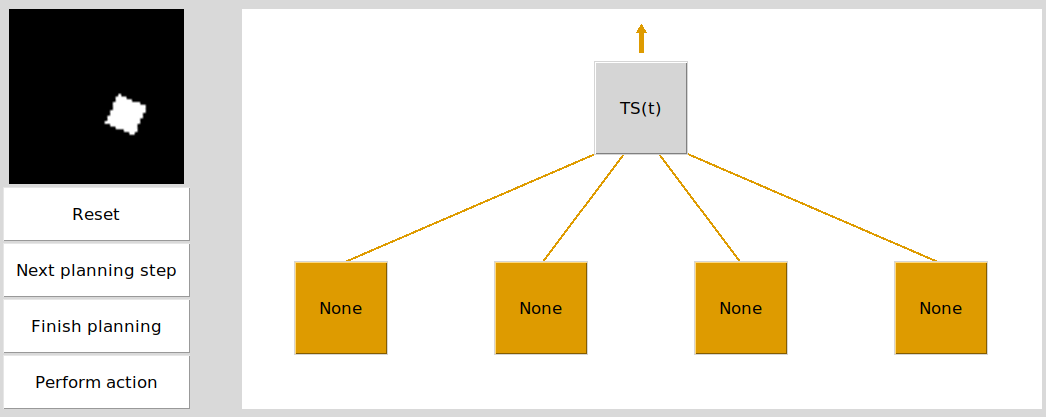}
	\end{center}
  \caption{This figure illustrates the visualisation frame of the GUI used to analyse a $BTAI_{3MF}$ agent. The image corresponding to the current state of the environment is displayed in the upper-left corner. Under the image are four buttons allowing the user to: reset the environment and agent, perform the next planning iteration, perform all the remaining planning iterations, and perform the current best action in the environment. Finally, on the right hand side of the image is a depiction of the MCTS planning, where $TS(t)$ represents the current temporal slice. At the moment, the current temporal slice has no children, and therefore its children are displayed in orange with the text ``None". Additionally, the current slice has no parent because it is the tree's root. Therefore, the arrow above the $TS(t)$ node is also orange.}
   \label{fig:gui_vf}
\end{figure}

\begin{figure}[H]
	\begin{center}
	\includegraphics[scale=0.4]{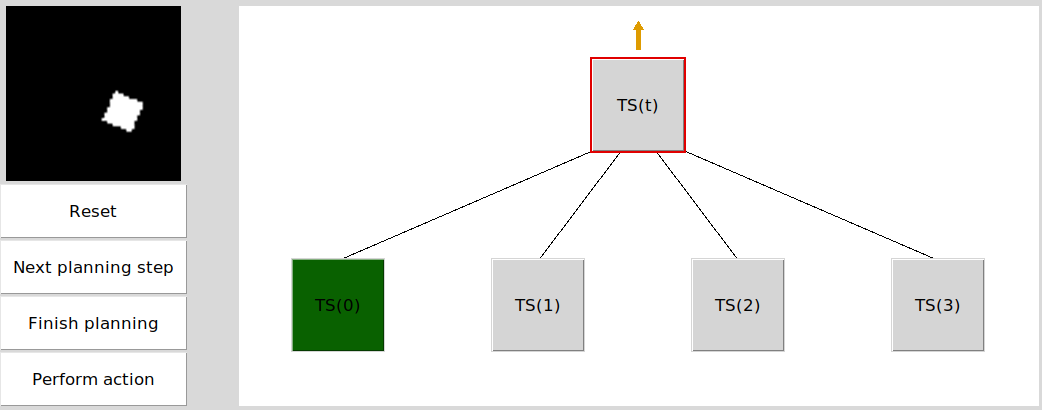}
	\end{center}
  \caption{This figure illustrates the visualisation frame of the GUI used to analyse a $BTAI_{3MF}$ agent after performing one planning iteration. The children of the root node are now available. One of them is displayed in green, it corresponds to the best action found so far by the MCTS algorithm. The root node has a red square surronding it, which means that it was selected for expansion by the MCTS algorithm.}
   \label{fig:gui_vf_next_planning_step}
\end{figure}

\begin{figure}[H]
	\begin{center}
	\includegraphics[scale=0.3]{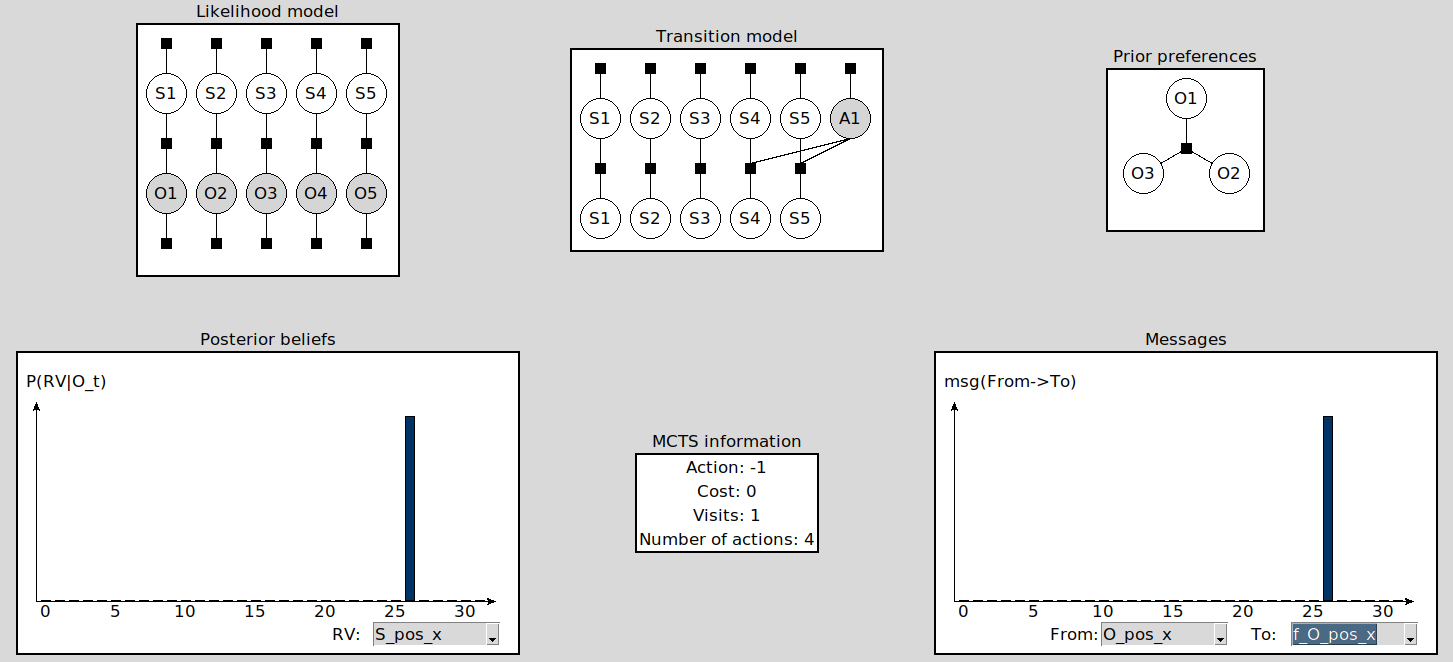}
	\end{center}
  \caption{This figure illustrates the frame displaying the information of the current temporal slice of the $BTAI_{3MF}$ agent. Six widgets are displayed. The first displays the structure of the likelihood model using the factor graph formalism. On this graph, we see that the model is composed of five obervations and five hidden states. Each observation depends on only one hidden state. The second widget displays the structure of the transition mapping. We see that only two hidden states depend on the action taken by the agent, i.e., the hidden states corresponding to the X and Y position of the shape. The third widget shows the structure of the prior preferences. Here, there is only one factor over three random variables, i.e., the shape and its (X, Y) position. Note, when moving your mouse over a variable in the likelihood, transition or prior preference widget the complete name of the variable is displayed, e.g., when moving over ``S1" the label ``S\_shape" is displayed. The fourth widget illustrates the posterior over the latent variable corresponding to the x position of the shape. The random variable whose posterior is displayed can be changed either by using the combo box in the bottom-right corner of the widget or by clicking on a latent variable in the likelihood model widget. The fifth widget displays information related to the Monte-Carlo tree search. Finally, the last widget illustrates the message sent from the observation variable corresponding to the X position of the shape to its likelihood factor.}
   \label{fig:gui_tsf_for_current_ts}
\end{figure}

\begin{figure}[H]
	\begin{center}
	\includegraphics[scale=0.4]{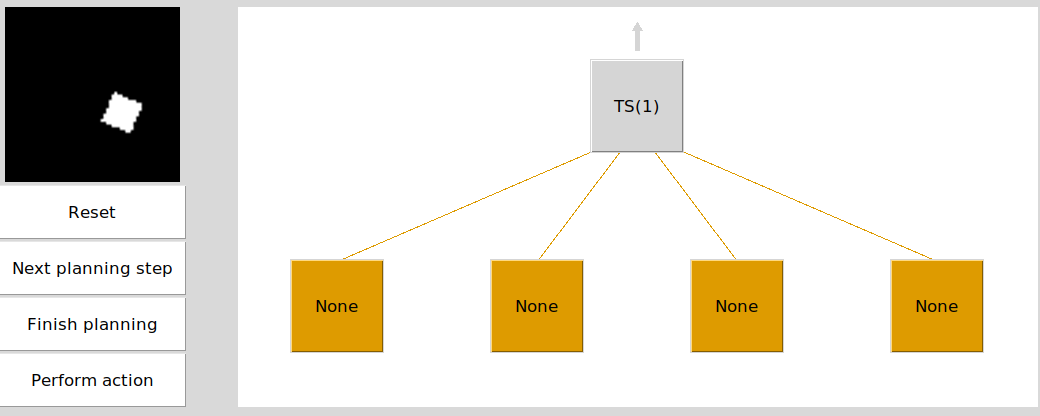}
	\end{center}
  \caption{This figure illustrates what happens when clicking on the child ``TS(1)" in Figure \ref{fig:gui_vf_next_planning_step}. Put simply, ``TS(1)" becomes the new root and we see that its children have not been expanded yet. Additionally, the arrow above the ``TS(1)" node is gray meaning that this node has a parent, i.e., ``TS(t)". Clicking on this arrow leads us back to Figure \ref{fig:gui_vf_next_planning_step}.}
   \label{fig:navigating_to_child}
\end{figure}

\begin{figure}[H]
	\begin{center}
	\includegraphics[scale=0.3]{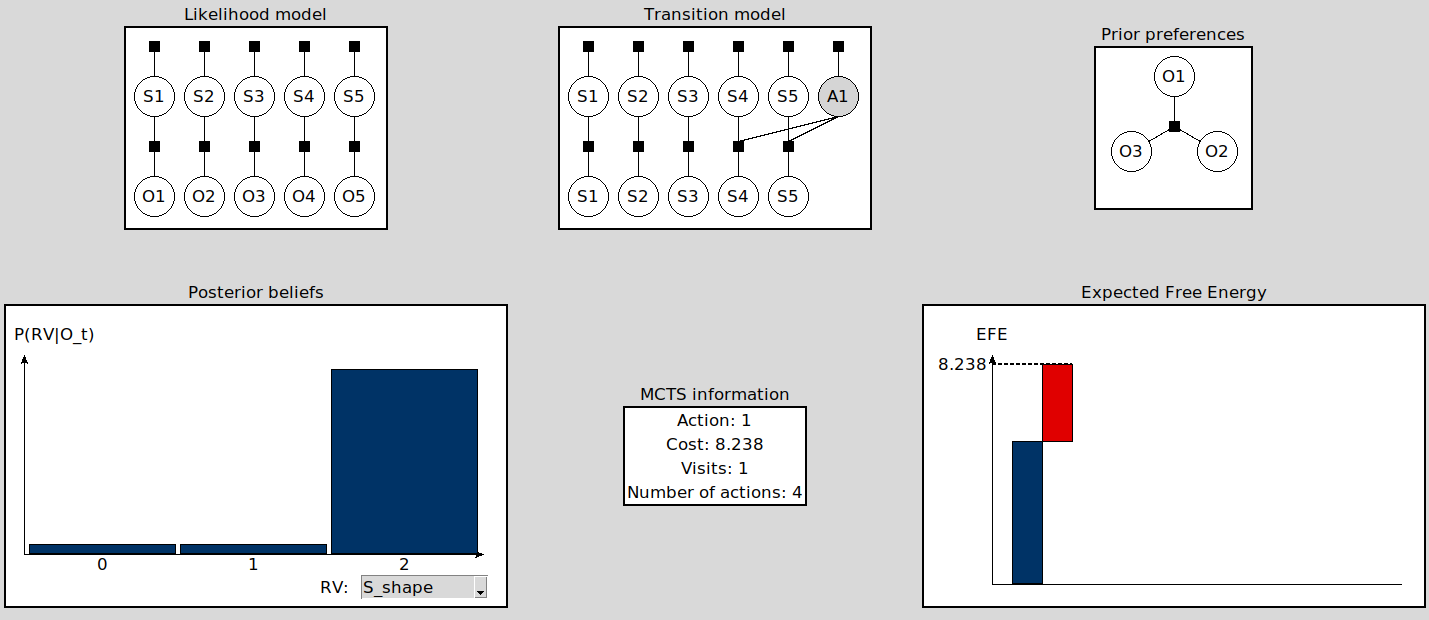}
	\end{center}
  \caption{This figure illustrates what happens when clicking on ``TS(1)" in Figure \ref{fig:navigating_to_child}. Most of the widgets have already been explained with the exception of the one in the bottom right-corner, which displays how the expected free energy decomposes into risk (blue box) and ambiguity (red box). When clicking on the blue or red box, the decomposition of the risk or ambiguity term is displayed as shown in Figure \ref{fig:ambiguity_decomposition}.}
   \label{fig:ts_frame_in_the_future}
\end{figure}

\begin{figure}[H]
	\begin{center}
	\includegraphics[scale=0.5]{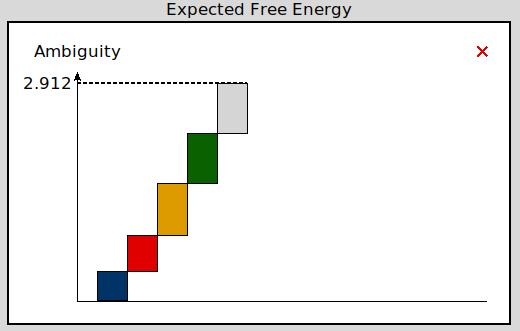}
	\end{center}
  \caption{This figure illustrates how the ambiguity term decomposes into the ambiguity of the likelihood of each observed variable, i.e., the ambiguity of ``O\_shape" in blue, ``O\_scale" in red, ``O\_orientation" in orange, ``O\_pos\_x" in green, and ``O\_pos\_y" in gray.}
   \label{fig:ambiguity_decomposition}
\end{figure}

In this appendix, we describe how to analyse a $BTAI_{3MF}$ agent using our graphical user interface (GUI). The relevant code can be found in the file \url{analysis_BTAI_3MF.py} at the following URL: \url{https://github.com/ChampiB/BTAI_3MF}. The first step is to create the environment and agent as described in Appendix A. Then, we create a GUI object and run the main loop as follows:
\begin{Verbatim}[commandchars=\\\{\}]
\PYG{c+c1}{\PYGZsh{} Create the GUI for analysis.}
\PYG{n}{gui} \PYG{o}{=} \PYG{n}{GUI}\PYG{p}{(}\PYG{n}{env}\PYG{p}{,} \PYG{n}{agent}\PYG{p}{)}
\PYG{n}{gui}\PYG{o}{.}\PYG{n}{loop}\PYG{p}{()}
\end{Verbatim}
The above two lines should open a graphical user interface as shown in Figure \ref{fig:gui_vf}. When clicking on the node of the current temporal slice $TS(t)$, one can obtain additional information about this temporal slice, c.f., Figure \ref{fig:gui_tsf_for_current_ts}. When clicking on the button named ``Next planning iteration" in Figure \ref{fig:gui_vf}, a planning iteration is performed and the tree displayed on the right-hand-side of this frame is updated as shown in Figure \ref{fig:gui_vf_next_planning_step}. When clicking on the root's children, e.g., ``TS(1)", it is possible to navigate through the tree created by the MCTS algorithm as shown in Figure \ref{fig:navigating_to_child}. When ``TS(1)" is displayed as the new root as in Figure \ref{fig:navigating_to_child}, clicking on ``TS(1)" again will display the information of this node as depicted by Figure \ref{fig:ts_frame_in_the_future}. Finally, Figure \ref{fig:ambiguity_decomposition} shows how the ambiguity term of the expected free energy can be decomposed into its component parts.

\section*{Appendix C: sum-rule, product-rule and d-separation criterion.}

In this appendix, we explain three important properties than are used in the core of the paper, namely: the sum-rule and product-rule of probability and the d-separation criterion.

\subsection*{Sum-rule of probability}

Given a set of random variables $X = \{X_1, ..., X_n\}$, and a joint distribution $P(X_1, ..., X_n)$ over $X$. The sum-rule allows to sum out a subset of the random variables. Here are a few examples:
\begin{align*}
P(X_1, ..., X_{n-1}) &= \sum_{X_n} P(X_1, ..., X_n),\\
P(X_1, ..., X_{n-2}) &= \sum_{X_{n-1}} \sum_{X_n} P(X_1, ..., X_n),\\
P(X_1, ..., X_{n-3}) &= \sum_{X_{n-2}} \sum_{X_{n-1}} \sum_{X_n} P(X_1, ..., X_n).
\end{align*}
Note, the sum-rule can also be used with a conditional distribution $P(X_1, ..., X_n|Y_1, ..., Y_m)$, for examples:
\begin{align*}
P(X_1, ..., X_{n-1}|Y_1, ..., Y_m) &= \sum_{X_n} P(X_1, ..., X_n|Y_1, ..., Y_m),\\
P(X_1, ..., X_{n-2}|Y_1, ..., Y_m) &= \sum_{X_{n-1}} \sum_{X_n} P(X_1, ..., X_n|Y_1, ..., Y_m),\\
P(X_1, ..., X_{n-3}|Y_1, ..., Y_m) &= \sum_{X_{n-2}} \sum_{X_{n-1}} \sum_{X_n} P(X_1, ..., X_n|Y_1, ..., Y_m).
\end{align*}

\subsection*{Product-rule of probability}

Given a set of random variables $X = \{X_0, ..., X_n\}$, and a joint distribution $P(X_0, ..., X_n)$ over $X$. The product-rule allows us to factorise the joint into a product of factors without doing any conditional independence assumptions about $P(X_1, ..., X_n)$. More formally:
\begin{align*}
P(X_0, ..., X_n) &= P(X_n)\prod_{i = 0}^{n-1} P(X_i|X_{i+1:n}),
\end{align*}
where $X_{i:j} = \{X_i, ..., X_j\}$ is the set of random variables containing all the variables between $X_i$ and $X_j$ (included). Note, the product-rule can also be used with a conditional distribution $P(X_0, ..., X_n|Y_1, ..., Y_m)$:
\begin{align*}
P(X_0, ..., X_n|Y_1, ..., Y_m) &= P(X_n|Y_1, ..., Y_m)\prod_{i = 0}^{n-1} P(X_i|X_{i+1:n}, Y_1, ..., Y_m).
\end{align*}

\subsection*{The d-separation criterion}

The d-separation criterion is a tool than can be used to check whether two sets of random variables ($X$ and $Y$) are independent given a third set of random variables $Z$. More formally, the d-separation criterion is a tool to check whether $X \indep Y\,| \,Z$. Knowing that $X \indep Y\,|\,Z$ holds in a distribution $P$ is useful because if $X \indep Y\,|\,Z$, then:
\begin{align*}
P(X, Y|Z) &= P(X| Y, Z)P(Y|Z) \tag{product-rule}\\
&= P(X|Z)P(Y|Z).\tag{$X \indep Y \,|\, Z$}
\end{align*}
First, let $G = (\mathcal{X}, \mathcal{E})$ be a graph over a set of nodes $\mathcal{X}$ connected by a set of directed edges $\mathcal{E}$. Given two nodes in the graph (i.e., $N_i, N_j \in \mathcal{X}$), we note: (i) $N_i \rightarrow N_j$ if there is a directed edge from $N_i$ to $N_j$ in the graph, (ii) $N_i \leftarrow N_j$ if the graph contains a directed edge from $N_j$ to $N_i$, and (iii) $N_i \rightleftarrows N_j$ if (i) or (ii) holds. Second, we say that there is a \textit{trail} between two nodes (i.e., $N_1, N_n$) in the graph, if there is a sequence of distinct nodes $N = (N_1, ..., N_n)$, such that: $N_i \rightleftarrows N_{i+1}$ holds for all $i \in \{1, ..., n-1\}$. Third, we say that a trail between $N_1$ and $N_n$ is \textit{active} if: (a) each time there is a v-structure (i.e., $N_{i-1} \rightarrow N_i \leftarrow N_{i+1}$) in the trail, then either $N_i$ or (at least) one of its descendants are in $Z$, and (b) no other node along the trail are in $Z$. Finally, we say that $X$ and $Y$ are \textit{d-separated} by $Z$ if for all $X_i \in X$ and $Y_i \in Y$ there is no active trail between $X_i$ and $Y_i$ (given $Z$).

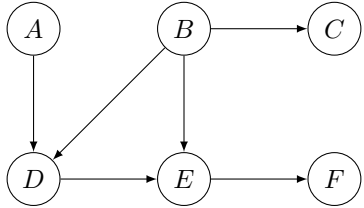
\begin{figure}[H]
	\begin{center}
	\begin{tikzpicture}[square/.style={regular polygon,regular polygon sides=4}]
		\node[latent] (A) at (0,0.5) {$A$};
		\node[latent] (B) at (2,0.5) {$B$};
		\node[latent] (C) at (4,0.5) {$C$};
		\node[latent] (D) at (0,-1.5) {$D$};
		\node[latent] (E) at (2,-1.5) {$E$};
		\node[latent] (F) at (4,-1.5) {$F$};
        \draw [-latex] (A) -- (D);
		\draw [-latex] (D) -- (E);
		\draw [-latex] (E) -- (F);
		\draw [-latex] (B) -- (D);
		\draw [-latex] (B) -- (E);
		\draw [-latex] (B) -- (C);
    \end{tikzpicture}
 	\end{center}
\vspace{-0.25cm}
    \caption{
This figure illustrates a Bayesian network in which the following independences assumptions \textbf{hold}: $A \indep B\,|\, \emptyset$; $A, D \indep C\,|\,B$; and $A \indep E\,|\,D, B, C$. In contrast, the following independences assumptions \textbf{does not hold}: $A \indep B\,|\, D$; $A \indep E\,|\,B, C$; and $A \indep B\,|\,E$ .
}
    \label{fig:d_sep_BN}
\end{figure}

Using our terminology, the d-separation criterion states that if $X$ and $Y$ are d-separated by $Z$ in a graph $G$ representing the factorisation of a distribution $P$, then $X \indep Y\,|\,Z$ holds in the distribution $P$. Intuitively, the d-separation criterion help us to determine whether $X \indep Y\,|\,Z$ holds in $P$ by looking at the topology of the graph $G$. For example, consider the Bayesian network illustrated in Figure \ref{fig:d_sep_BN}, and let $P$ be the joint distribution represented by this Bayesian network. Using the product rule, we get:
\begin{align*}
P(A, B, C, D, E, F) &= P(F|A, B, C, D, E)P(E|A, B, C, D)P(C|A, B, D)P(D|A, B)P(B|A)P(A).
\end{align*}
Note, that all trails between $C$ and $A, D$ are blocked by $B$, i.e., there is no active trails between $C$ and $A, D$ given $B$. Thus, we have $C \indep A, D\,|\,B$ and:
\begin{align*}
P(A, B, C, D, E, F) &= P(F|A, B, C, D, E)P(E|A, B, C, D)\bm{P(C|B)}P(D|A, B)P(B|A)P(A).
\end{align*}
Moreover, there is no active trail between $B$ and $A$ given $\emptyset$, therefore $B \indep A\,|\,\emptyset$ and:
\begin{align*}
P(A, B, C, D, E, F) &= P(F|A, B, C, D, E)P(E|A, B, C, D)P(C|B)P(D|A, B)\bm{P(B)}P(A).
\end{align*}
Using the same reasoning, one can see that $F \indep A, B, C, D\,|\,E$ and thus:
\begin{align*}
P(A, B, C, D, E, F) &= \bm{P(F|E)}P(E|A, B, C, D)P(C|B)P(D|A, B)P(B)P(A).
\end{align*}
Finally, using the d-separation one more time leads to the following factorisation for $P$:
\begin{align*}
P(A, B, C, D, E, F) &= P(F|E)\bm{P(E|B, D)}P(C|B)P(D|A, B)P(B)P(A).
\end{align*}

\end{document}